\documentclass{article}

\PassOptionsToPackage{numbers, sort&compress}{natbib}

\usepackage[preprint]{neurips_2021}

\usepackage[utf8]{inputenc} 
\usepackage[T1]{fontenc}    
\usepackage{hyperref}       
\usepackage{url}            
\usepackage{booktabs}       
\usepackage{amsfonts}       
\usepackage{nicefrac}       
\usepackage{microtype}      
\usepackage{xcolor}         
\usepackage{natbib}

\usepackage{graphicx}
\usepackage{mathtools}
\usepackage{bm}
\usepackage{siunitx}
\usepackage{tabularx}
\usepackage{multirow}
\usepackage{booktabs}
\usepackage{tablefootnote}
\usepackage{longtable}
\usepackage{threeparttable}
\usepackage{wrapfig}

\hypersetup{
bookmarksopen=true,
bookmarksnumbered=true,
unicode=false,
pdftoolbar=true,
pdfmenubar=true,
pdffitwindow=false,
pdfstartview={FitH},
pdftitle={My title},
pdfauthor={Author},
pdfsubject={Subject},
pdfcreator={Creator},
pdfproducer={Producer},
pdfkeywords={keywords},
pdfnewwindow=true,
colorlinks=true,
linkcolor=blue,
citecolor=blue,
filecolor=magenta,
urlcolor=blue}

\title{Uncovering Closed-form Governing Equations of Nonlinear Dynamics from Videos}

\author{Lele Luan, Yang Liu \& Hao Sun\thanks{Corresponding author}  \\
Northeastern University, Boston, MA, USA \\
\texttt{\{luan.l, yang1.liu, h.sun\}@northeastern.edu}
}

\begin{document}

\maketitle

\begin{abstract}
  Distilling analytical models from data has the potential to advance our understanding and prediction of nonlinear dynamics. Although discovery of governing equations based on observed system states (e.g., trajectory time series) has revealed success in a wide range of nonlinear dynamics, uncovering the closed-form equations directly from raw videos still remains an open challenge. To this end, we introduce a novel end-to-end unsupervised deep learning framework to uncover the mathematical structure of equations that governs the dynamics of moving objects in videos. Such an architecture consists of (1) an encoder-decoder network that learns low-dimensional spatial/pixel coordinates of the moving object, (2) a learnable Spatial-Physical Transformation component that creates mapping between the extracted spatial/pixel coordinates and the latent physical states of dynamics, and (3) a numerical integrator-based sparse regression module that uncovers the parsimonious closed-form governing equations of learned physical states and, meanwhile, serves as a constraint to the autoencoder. The efficacy of the proposed method is demonstrated by uncovering the governing equations of a variety of nonlinear dynamical systems depicted by moving objects in videos. The resulting computational framework enables discovery of parsimonious interpretable model in a flexible and accessible sensing environment where only videos are available.
\end{abstract}

\section{Introduction} \label{introduction}

In many scientific disciplines, discovering governing equations (e.g., PDEs, ODEs) allows for understanding, modeling and prediction of complex dynamical systems. Increasing richness of data and advances in machine learning have given rise to a new paradigm of understanding unknown physical systems called data-driven governing equation discovery \cite{rudy2017data,champion2019data}. Since video camera has been widely used in scientific data collection, discovering potential physical laws or governing equations from video-format data, a subset of dynamical scene understanding, has led to expanded interest in the computer vision community \cite{chen2021grounding}. Advances in deep learning, especially convolutional neural networks (CNNs), also has led to a major impact on the current state-of-the-art scene understanding from videos.

Physical scene understanding from videos via deep learning started with object property learning or simple law discovery \cite{wu2016physics,fragkiadaki2016learning}. For example, a generative model was proposed to determine the parameters controlling the dynamical systems by incorporating with known physical engines \cite{wu2015galileo}. 
In most deep learning approaches, explicit physical law is distilled after object moving trajectory being determined by supervised regression from videos \cite{watters2017visual,de2018end}. Later, unsupervised learning approaches for object location \cite{kosiorek2018sequential,hsieh2018learning} promoted the parameter estimation of dynamical system from videos in an unsupervised scheme \cite{jaques2020physics}. Noteworthy, since the object is localized in the image space, these approaches can only learn physical laws or their parameters established in spatial (pixel) coordinate such as billiards, gravity, spring, bouncing, etc \cite{jaques2020physics,de2020discovery,kossen2019structured,purushwalkam2019bounce}. The nonconformity between the spatial coordinates of moving object in the image space and the real physical states (where the physical law can be explicitly expressed) limits the applicability of these approaches to the discovery of more complex dynamical systems. Furthermore, most of existing methods focus on the identification of equation parameters in which the physical law structure is known or at least partially known. In addition, high roughness of moving trajectory extracted by current networks restricts the closed-form governing equation discovery which typically requires clear (or low-level noisy) state data.

\begin{wrapfigure}[11]{r}{0.7\textwidth}
\vspace{-16pt}
\begin{center}
\includegraphics[width=1\linewidth]{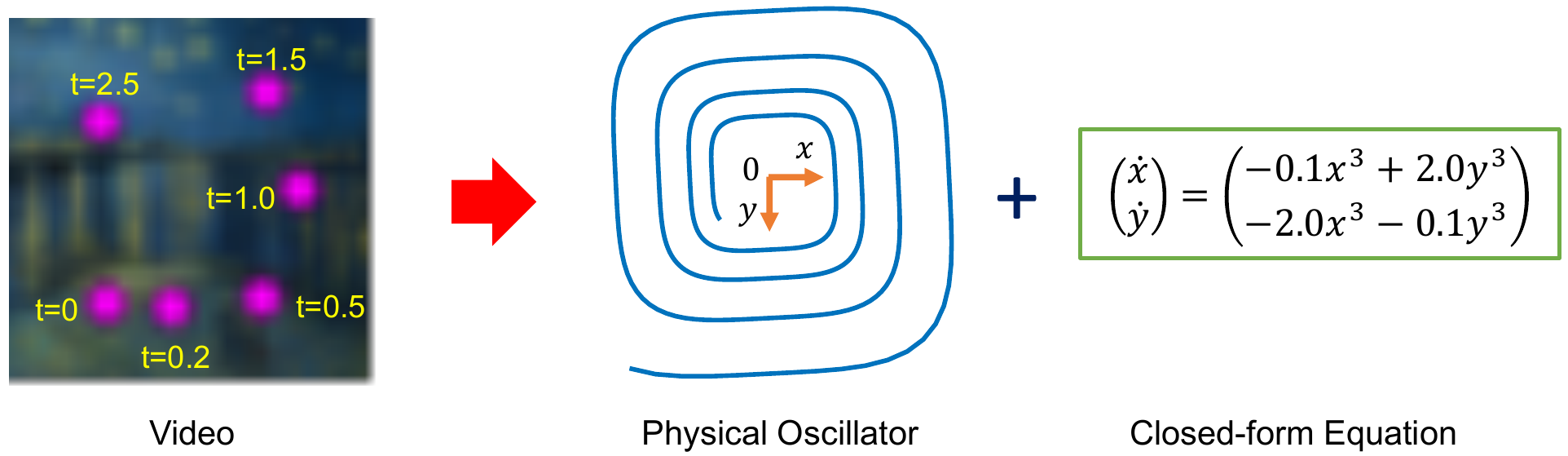}
\vspace{-12pt}
\caption{Simultaneously extracting the physical states of moving object and uncovering its closed-form governing equations from video.}
\label{fig:target}
\end{center}
\vspace{-6pt}
\end{wrapfigure}
\normalsize

Although great success has been recently achieved on discovering governing equations of nonlinear dynamics from directly measured system states (e.g., trajectory time series) \cite{brunton2016discovering,long2019pde,udrescu2020ai}, identifying closed-form governing equations straightly from raw videos is still a grand challenging. In this work, we propose a novel end-to-end unsupervised deep learning framework to uncover the closed-form explicit physical laws of nonlinear dynamics directly from videos. The task we intend to resolve shown in Fig. \ref{fig:target} demonstrates the paradigm we build seeking to simultaneously extract the physical states of moving object in the scene and uncover their governed equation. Unlike existing methods which discover physical law from spatial/pixel coordinate trajectories of moving object, our method uncovers the explicit governing equations for physical states instead, which makes it possible to represent more complex dynamical systems. Furthermore, the physical states are not extracted independently from the video but under the constraint of the underlying physical law. The joint optimization not only helps the extraction of physical states, but also leads to the identification of closed-form governing equations.

\subsection{Contribution} \label{contribution}
Our key contribution lies in the innovative solution to uncover closed-form governing equations directly from videos without knowing the physical law structure, based on a novel end-to-end learning architecture. This framework seamlessly combines the latest unsupervised object localization, spatial-physical coordinate transformation and sparse regression on physical sates, to achieve successful discovery. We demonstrate the capacity of the method on discovering nonlinear dynamics of moving object in videos governed by a series of 1st and 2nd order ordinary differential equation (ODEs). 

\subsection{Related Work} \label{related_work}
Data-driven governing equation discovery started with building mathematical models for dynamical systems from given measurement of physical states. Followed by early methods that identified linear models from input-output data \cite{schmid2010dynamic,kutz2016dynamic}, increasing interests have been placed on learning interpretable nonlinear models from data. In particular, sparse regression techniques, which rely on a library of possible candidates and prune redundant terms to keep fewest ones necessary to balance the complexity with descriptive ability, has  achieved great success on identifying exact expressions for nonlinear ODEs \cite{brunton2016discovering,schaeffer2017sparse,mangan2017model,loiseau2018constrained,schaeffer2018extracting,thaler2019sparse,schaeffer2020extracting}, PDEs \cite{rudy2017data,schaeffer2017learning,narasingam2018data,gurevich2019robust,reinbold2019data,rudy2019data,mangan2019model,maslyaev2019data,chen2020deep} and stochastic differential equations \cite{boninsegna2018sparse,li2021data}. In particular, the sparse identification of nonlinear dynamics (SINDy) \cite{brunton2016discovering} approach especially has led to variant algorithms to identify ODEs for a wide range of nonlinear dynamical systems \cite{mangan2016inferring,loiseau2018sparse,kaiser2018sparse,quade2018sparse,champion2019discovery,champion2019data,li2019discovering,lai2019sparse,kaheman2020sindy,cichos2020machine}. Later, the compromise of accuracy and robustness of sparse regression due to the challenges of computing derivatives of noisy measurement data promoted the development of algorithms to produce accurate derivative approximations \cite{van2020numerical,sun2021physics}, as well as the de-noising of data and identifying the underlying parsimonious dynamics simultaneously \cite{rudy2019deep,kaheman2020automatic}.

Instead of discovering physical laws from given state time series, there has been increased interest in physical scene understanding from video in recent years. Physical scene understanding started with learning ``blur'' physics, in which physical laws are not expressed explicitly but simulated by network physical modules. Inspired by powerful feature extraction ability, deep neural network based methods were leveraged to build data-driven models of intuitive physics at first \cite{fragkiadaki2016learning,bhattacharyya2016long,ehrhardt2017learning,finn2016unsupervised}. Some interaction models were also developed to consider the interactions between entities in the scene \cite{watters2017visual,battaglia2018relational}. Later, the physical engines/modules were proposed and incorporated with the extracted object-based representations or physical states to develop the video description or prediction \cite{purushwalkam2019bounce,chang2017compositional,wu2017learning,zheng2018unsupervised,janner2019reasoning,ye2018interpretable,fraccaro2017disentangled,de2018end,van2018relational}. Although some of those approaches considered interpretable parameters in the physical modules like object mass, position, speed and friction \cite{wu2017learning,chang2017compositional,ye2018interpretable}, the physical laws are still not explicitly discovered and the physics are condensed as state space, or simulated as neural physics engines/modules to better predict the temporal video state.

In order to improve the interpretability of the discovered physical law, learning explicit dynamics (governing equations or physical properties) has recently become popular in physical scene understanding. Several hybrid methods take a data-driven approach to estimate real mechanical process from video sequences \cite{wu2015galileo,wu2016physics}, or model Newtonian physics via latent variable to predict motion trajectories in images \cite{mottaghi2016newtonian}. Since explicit physical dynamics learning from videos requires to extract the motion and model the object dynamics in the scene, the Two-step discovery has become the most common strategy, where the physical law is identified after the object moving trajectory being extracted \cite{stewart2017label,wu2017learning,belbute2018end, ehrhardt2018unsupervised}. Later, advances in unsupervised object localization like spatial transformers (ST) \cite{kosiorek2018sequential,hsieh2018learning,ehrhardt2018unsupervised,zhu2018object}, Position-Velocity Encoders (PVEs) \cite{jonschkowski2017pves} enabled the explicit physical law discovery in an unsupervised scheme \cite{kossen2019structured,jaques2020physics}. However, these approaches require a strong \textit{a prior} knowledge on the structure of the physical law or governing equation (e.g., the equation form is assumed known while the coefficients might be unknown). Furthermore, these methods rely on the pixel coordinates of moving object which restrict the discovery of closed-form equations for complex dynamical systems where the physical states cannot be simply described (or treated equal to) the pixel coordinates. 

Recently, there has been attempt to uncover closed-form equations of low-dimensional representations from videos or high-dimensional data. Champion et al. \cite{champion2019data} proposed an autoencoder network to condense high-dimensional data into a reduced space where the dynamics may be sparsely represented, and the governing equations and associated coordinate system are learned simultaneously. However, this approach was designed to discover governing equations from high-dimensional data where the derivatives of latent physical variables can be calculated by propagating the derivative of input through the network. Our work is also partially inspired by the unsupervised method for physical parameter estimation from video proposed in \cite{jaques2020physics}, which, however, can only discover the parameters of given physical laws which are established in the context of pixel coordinate. Although Udrescu and Tegmark \cite{udrescu2020symbolic} provided a method for closed-form governing equation discovery from videos, this two-step method is different with our proposed end-to-end scheme. To the best of our knowledge, our model is the first to offer an end-to-end unsupervised closed-form governing equation discovery directly from videos.

\section{Method} \label{method}

We present an end-to-end unsupervised deep learning scheme to uncover the closed-form governing equations from raw videos in which the dynamics is represented by a moving object. In this method, an encoder-decoder is employed to extract the spatial coordinate of moving object which is then mapped into physical states (or physical coordinate) by a trainable spatial-physical coordinate transformer under the constraint of physical law. The closed-form governing equations for the dynamics will be in reverse uncovered from the extracted physical states by sparse regression. The designed network architecture and training procedure are given as follows.   

\begin{figure}[t!]
	\centering
	\includegraphics[width=0.99\linewidth]{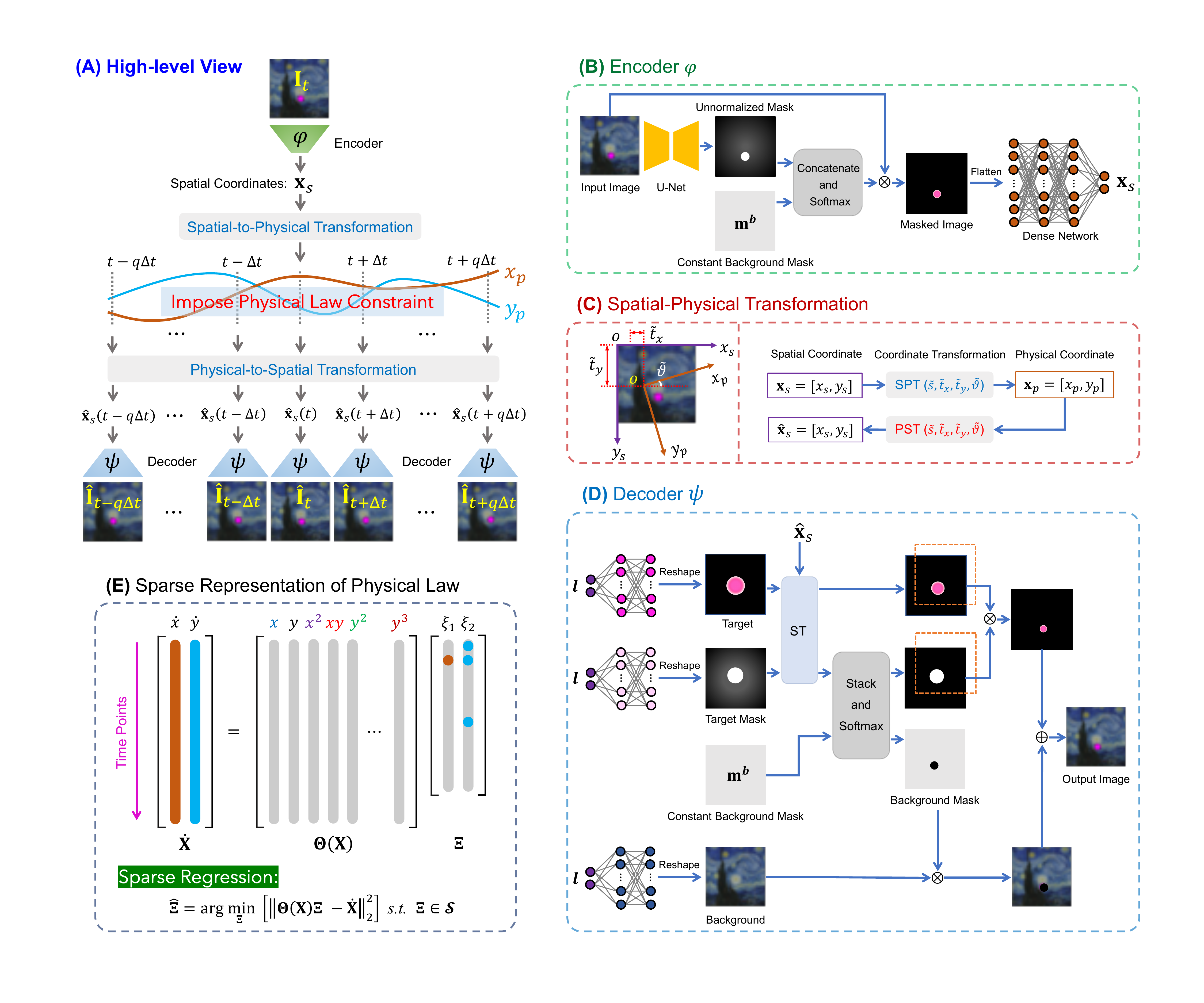}
	\caption{Schematic architecture of the proposed end-to-end unsupervised deep learning with sparse regression to simultaneously extract physical states of the dynamic and uncover the underlying closed-form governing equations from videos. (A) represents the high-level view of the designed architecture. The time-dependent physical states \(x_p\) and \(y_p\) extracted from autoencoder has to follow a set of parsimonious underlying governing equations, which in turn constrain the autoencoder. The single video frame \(\mathbf{I}_t\) is feed into (B) Encoder \(\varphi\) to capture the 2D spatial coordinate of the moving object, which is then mapped into physical states by (C) Spatial-Physical Transformation (SPT) where the position of moving object is kept while the coordinate system is transformed via the learnable transformation factors including scaling \(\tilde{s}\), translation \(\tilde{\mathbf{t}}=(\tilde{t}_x,\tilde{t}_y)^T\) and rotation \(\tilde{\vartheta}\). The physical states then enters the Physical-Spatial Transformation (PST), with the same transformation factors as SPT, to reproduce the spatial coordinate, which is followed by (D) Coordinate-Consistent Decoder \(\psi\) to reconstruct the video frame with \(\mathbf{\hat{I}}_t\). The extracted physical states are described by a library of candidate functions \(\mathbf{\Theta}(\mathbf{X})\) parameterized by the unknown sparse coefficients \(\mathbf{\Xi}\in\boldsymbol{\mathcal{S}}\), where $\boldsymbol{\mathcal{S}}$ is a sparsity constraint set. The physical states at frame \(\mathbf{I}_t\) are integrated numerically to simulate the forward and backward dynamics evolution in time \(q\)-steps with time step \(\Delta{t}\). The integrated physical states are passed into PST to recover the spatial coordinates of moving object at neighbor snapshots and reconstruct those frames. In the Encoder and Decoder, \(\mathbf{m}^b\), a tensor of all ones, represents the constant mask for background before being passed into Softmax. In the Decoder, \(l\) denotes a constant array of 1s, serving as the input of fully-connected networks to learn the object, object mask and background, respectively. More details about the network architecture is given in Appendix \ref{network_architecture_appendix}.}
	\label{fig:network_architecture}
	\vspace{-12pt}
\end{figure}

\paragraph{Network Architecture.}
In order to learn the physical equations from videos, several components are considered in place. First of all, the high-dimensional image is condensed into low-dimensional latent variables, the spatial coordinate of moving object by an encoder. Conversely, a decoder reconstructs the video frames from the represented low-dimensional latent variables. Since the latent spatial coordinate from the encoder-decoder has gap with the physical states which are used to represent the underlying physical law, a trainable spatial-physical coordinate transformation is added to build the inter-conversion between spatial and physical coordinates of the moving object. In addition, because the dynamics to be discovered involves temporal evolution, consecutive video snapshots are considered for temporal integration (time marching) of the physical states. We assume that the governing equations that assemble the physical states only consist of a few important terms. Since the equation structure is unknown, we leverage a library, composed of a finite number of pre-defined possible linear/nonlinear candidate terms in the context of physical states extracted from the autoencoder, to construct the equations parameterized by unknown coefficients where are sparse in the solution space. This enables the temporal integration of physical states in consecutive video snapshots. The designed architecture is depicted in Fig. \ref{fig:network_architecture}, where Fig. \ref{fig:network_architecture}(A) shows the high-level view of the network architecture involving temporal integral of the physical states. 

\underline{Encoder}. The Encoder, a two-stage localization network proposed in \cite{jaques2020physics}, takes a single video frame \(\mathbf{I}_t\) as input and outputs a vector \(\mathbf{x}_s=(x_s,y_s)\) corresponding to the 2D spatial coordinate of the moving object. First, the input frame is passed through a U-Net \cite{ronneberger2015u} to produce an unnormalized mask. The unnormalized mask is concatenated with a constant background mask \(\mathbf{m}^b\) and then passed through a Softmax to produce the mask for the moving object and background. The multiplication between input and masks are fed into a fully-connected network to obtain the spatial coordinate of the moving object. The output layer of the encoder depends on the number of coordinates needed to describe the object location and the size of image. For a 2D physical system with image size of \(H \times H\), the latent space has two variables and the activation of output layer has a saturating non-linearity \(H/2\cdot\tanh{(\cdot)}+H/2\), which leads to the spatial coordinate of the moving object \(\mathbf{x}_s\) with values in \([0, H]\).

\underline{Coordinate-Consistent Decoder}. The decoder takes as input the spatial coordinates given by the physical-spatial coordinate transformation of physical states, and outputs a reconstructed video frame \(\Tilde{\mathbf{I}_t}\). Since the spatial coordinate of moving object has specific relationship with entire image intensities when background is constant, here the Coordinate-Consistent Decoder proposed in \cite{jaques2020physics} is employed as decoder. In the Coordinate-Consistent Decoder, the decoded frame is reconstructed from the fixed correspondence between the spatial coordinate of moving object and pixel coordinate of image imposed by ST \cite{jaderberg2015spatial} with its reciprocal transformation parameters. The detailed illustration of the Coordinate-Consistent Decoder is given in Appendix \ref{network_architecture_appendix}.

\underline{Spatial-Physical Transformation}. The Encoder and Coordinate-Consistent Decoder introduced above are expected to extract the spatial coordinate of moving object. The mapping between the extracted spatial coordinate and physical state is built with coordinate transformation. If we assume the dynamics is present in the 2D plane of image space, the transformation between spatial and physical coordinates can be implemented by a standard 2D Cartesian coordinate transformation process (\(\mathcal{T}\) represents spatial to physical transformation, \(\tilde{\mathcal{T}}\) the inverse transformation). In coordinate transformation, the position of moving object is kept fixed while the original spatial (pixel) coordinate system is transformed relative to the object. With translation vector \(\tilde{\mathbf{t}}=(\tilde{t}_x,\tilde{t}_y)^T\), scaling factor \(\tilde{s}\) and rotation angle \(\tilde{\vartheta}\), the physical coordinate \(\mathbf{x}_p=(x_p,y_p)\) can be obtained from spatial coordinate by 
\(\mathbf{x}^T_p=\mathcal{T}(\mathbf{x}^T_s)=\tilde{s}\big[\mathbf{Q}(\tilde{\vartheta})\mathbf{x}^T_s-\tilde{\mathbf{t}}\big]\), where \(\mathbf{Q}(\tilde{\vartheta})\) is the transformation matrix induced by rotation angle \(\tilde{\vartheta}\). In the dataset studied here, since the real physical coordinate system is parallel to the spatial coordinate and the coordinate transformation has no rotation, the spatial to physical transformation is degraded to \(\mathbf{x}^T_p=\mathcal{T}(\mathbf{x}^T_s)=\tilde{s}(\mathbf{x}^T_s-\tilde{\mathbf{t}})\). Likewise, the physical to spatial transformation can be expressed as \(\mathbf{x}^T_s = \tilde{\mathcal{T}}(\mathbf{x}^T_p)=1/\tilde{s}\mathbf{x}^T_p+\tilde{\mathbf{t}}\). Here, \(\mathcal{T}\) and \(\tilde{\mathcal{T}}\) share the same parameters of \(\tilde{s}\) and \(\tilde{\mathbf{t}}\). Those parameters are trainable variables and will be optimized in the network training. With the Encoder, Coordinate-Consistent Decoder and coordinate transformers (\(\mathcal{T}\) and \( \tilde{\mathcal{T}}\)), a complete autoencoder is built for the condensation of video frames into physical latent states. The standard autoencoder loss function can be expressed as: \(\mathcal{L}_{recon} = \left\lVert\mathbf{I}_t-\psi(\tilde{\mathcal{T}}(\mathcal{T}(\varphi(\mathbf{I}_t))))\right\rVert_2^2\).

\underline{Physical Constraint and Governing Equation Discovery}. In the designed network, the constraint by the underlying physical law, represented by a set of nonlinear ODEs, is imposed to the physical states \(\mathbf{x}_p\) extracted from autoencoder. The closed-form governing ODEs will be distilled by sparse regression as part of the network architecture. Here, we consider the nonlinear dynamics of moving object whose governing equations can be expressed by \(\frac{d}{dt}\mathbf{x}_p(t)=\bm{f}(\mathbf{x}_p(t))\), where \(\mathbf{x}_p(t)=(x_p(t),y_p(t))^T\) denotes the physical states extracted from the autoencoder. We seek a parsimonious model for the dynamics, resulting in a function \(\bm{f}\) that contains only a few active terms. Here, the model discovery can be framed as a sparse regression problem. If the derivatives of the physical states are calculated, the snapshots are stacked to form data matrices \(\mathbf{X}=[\mathbf{x}_{p,1}, \mathbf{x}_{p,2}, \dotsc, \mathbf{x}_{p,m}]^T\in \mathbb{R}^{m\times 2}\) and \(\mathbf{\dot{X}}=[\dot{\mathbf{x}}_{p,1}, \dot{\mathbf{x}}_{p,2}, \dotsc, \dot{\mathbf{x}}_{p,m}]^T\in \mathbb{R}^{m\times 2}\) where $m$ is the number of data points of the physical states. Although \(\bm{f}\) is unknown, we can construct an extensive library of \(n\) candidate functions \(\mathbf{\Theta(\mathbf{\mathbf{X}})}=[\theta_1(\mathbf{X}), \theta_2(\mathbf{X}),  ..., \theta_n(\mathbf{X})] \in \mathbb{R}^{m \times n}\), where each \(\mathbf{\theta}_j\) denotes a candidate term. The candidate function library is used to formulate an over-determined system \(\dot{\mathbf{X}}=\mathbf{\Theta}(\mathbf{X})\mathbf{\Xi}\), where the unknown matrix \(\mathbf{\Xi}=[\bm{\xi}_1, \bm{\xi}_2,...,\bm{\xi}_n]\in \mathbb{R}^{n \times 2}\) is the set of coefficients that determine the active terms from \(\mathbf{\Theta(\mathbf{\mathbf{X}})}\) in the dynamics \(\bm{f}\). For the candidate function library, \(\theta_i(\mathbf{X})\) can be any candidate function that may describe the system dynamics \(\bm{f}(\mathbf{x}(t))\) such as polynomial functions \(\theta_i(\mathbf{X})=\mathbf{X}^3\). By solving the optimization, the model of system dynamics can be identified \(\frac{d}{dt}\mathbf{x}_p(t)=\bm{f}(\mathbf{x}_p(t)) \approx \mathbf{\Theta}(\mathbf{x}_p(t))\mathbf{\Xi}\). The coefficients \(\mathbf{\Xi}\) are learned concurrently with the neural network parameters as part of the training process. Many optimization techniques can be used to obtain the sparse coefficients, such as sequential thresholded least squares (STLSQ) \cite{brunton2016discovering,zhang2019convergence}, LASSO \cite{tibshirani1996regression}, sparse relaxed regularized regression (SR3) \cite{zheng2018unified,champion2020unified}, stepwise sparse regression (SSR) \cite{boninsegna2018sparse}, and Bayesian approaches \cite{pan2015sparse,zhang2018robust}. With the time derivative of physical states \(\mathbf{x}_p(t)\) being calculated by the central difference method, one can enforce accurate modeling of the dynamics by incorporating the following physical derivative term into the loss function: \(\mathcal{L}_{\dot{\mathbf{x}}_p} = \left\lVert\frac{d(\mathcal{T}(\varphi(\mathbf{I}_t))}{dt} - \mathbf{\Theta}((\mathcal{T}(\varphi(\mathbf{I}_t)))^T)\mathbf{\Xi}\right\rVert_2^2\).

As shown in Fig. \ref{fig:network_architecture}, the temporal evolution of extracted physical states also brings an additional physical constraint by using the estimated vector field of the system model, similar to the simultaneous de-noising and equation discovery method proposed in \cite{rudy2019deep,kaheman2020automatic}. With the true vector field \(f(\mathbf{x}_p(t))\) being estimated by \(\mathbf{\Theta}(\mathbf{x}_p(t))\mathbf{\Xi}\), integrating over a segment of time \(t_j\) to \(t_{j+1}\) gives the integrated vector field, or flow map \(\mathbf{x}_p(j+1)=\mathbf{F}(\mathbf{x}_p(j))=\mathbf{x}_p(j)+\int_{t_j}^{t_{j+1}}\mathbf{\Theta}(\mathbf{x}_p(\tau))\mathbf{\Xi}d\tau\). This can be generalized to integrate the system either forward or backward in time \(q\) steps, namely,  \(\mathbf{x}_p(j+q)=\mathbf{F}^q(\mathbf{x}_p(j))=\mathbf{x}_p(j)+\int_{t_j}^{t_{j+q}}\mathbf{\Theta}(\mathbf{x}_p(\tau))\mathbf{\Xi}d\tau\). To obtain \(\mathbf{x}_p(j+q)\), the 4th-order Runge-Kutta numerical method is employed to simulate the dynamics forward/backward propagation in time with \(q\)-steps for the physical states. Then the forward/backward video frames can be reconstructed via the decoder as \(\hat{I}_{j+q}=\psi(\tilde{\mathcal{T}}(\mathbf{x}_p(j+q)))\), which leads to the forward and backward frame reconstructions loss from the temporal integration of the physical states:
\begin{equation}
\mathcal{L}_{int}^q = \left\lVert{I}_{j+q}-\psi\big(\tilde{\mathcal{T}}\big(\mathbf{x}_p(j+q)\big)\big)\right\rVert_2^2=\left\lVert{I}_{j+q}-\psi\big(\tilde{\mathcal{T}}\big(\mathbf{x}_p(j)+\int_{t_j}^{t_{j+q}}\mathbf{\Theta}(\mathbf{x}_p(\tau))\mathbf{\Xi}d\tau\big)\big)\right\rVert_2^2
\end{equation}
We highlight that, although involvement of forward/backward propagation of physical states leads to multiple decoders in the network, the network is easy to converge because the object and background are learned with external networks and those decoders only have spatial coordinate as input. 

\paragraph{Loss Functions and Learning.} In addition to three loss terms given above, an \(\ell_{0.5}\) regularizer \(\mathcal{L}_{reg}\) on the sparse regression coefficients \(\mathbf{\Xi}\) is included, which promotes sparsity of the coefficients and therefore encourages a parsimonious model for the dynamics. The combination of 4 loss terms gives the overall loss function:
\begin{equation}
\label{loss}
\mathcal{L}_{total}=\mathcal{L}_{recon}+\lambda_1\mathcal{L}_{\dot{\mathbf{x}}_p}+\lambda_2\mathcal{L}_{int}+\lambda_3\mathcal{L}_{reg}
\end{equation}
where the hyperparameters \(\lambda_1\), \(\lambda_2\), \(\lambda_3\) determine the relative weighting of the 3 loss function terms. The explicitly defined loss function and discussion on weight choice are given in Appendix \ref{loss_function}.

The network is trained using a multi-step training strategy to succeed the discovery, which includes \((\mathbf{\romannumeral1})\) pre-training, \((\mathbf{\romannumeral2})\) total loss training, \((\mathbf{\romannumeral3})\) sequential thresholding, and \((\mathbf{\romannumeral4})\) refinement. In order to guarantee the learning of spatial coordinate of moving object, pre-training of the pure autoencoder (without the physical constraint) is conducted before optimizing the total loss function. In pre-training, only \(\mathcal{L}_{recon}\) is optimized and only variables for encoder and decoder are updated while other trainable variables being fixed. After pre-training, the pre-trained model is loaded for total loss optimization, which leads to the sparsity (many close-to-zero values) in candidate function coefficients \(\mathbf{\Xi}\). Then sequential thresholding is incorporated into the training procedure as a proxy for \(\ell_0\) sparsity to obtain a model with only a few active terms. In sequential thresholding, all coefficients below the threshold which determines the minimum magnitude for coefficient in the model are set to zero at fixed intervals through the training. Finally, after the redundant terms being filtered, the network is post-trained based on the identified equation structure. The network is trained by the Adam optimizer \cite{kingma2014adam}. In addition to the loss function weightings and candidate function coefficient threshold, the training requires the choice of several other hyperparameters including learning rate, order of discovered dynamic system, physical library functions and initial values of Spatial-Physical Transformation. Details of the training procedure and hyperparameter choice are discussed in Appendix \ref{choice_hyperparameters} and \ref{training_procedure}.

\section{Experiments and Analysis} \label{experiments_analysis}

We demonstrate the efficacy of our methodology on discovering a group of nonlinear dynamical systems from videos which include first order ODE systems like \textit{Duffing Oscillator}, \textit{Cubic Oscillator} and \textit{Van der Pol Oscillator}, and second order ODE systems including \textit{2D Oscillator}, \textit{Magnetic} and \textit{Quartic Oscillator}. The studied videos are generated by plotting simulated physical trajectories on randomly selected background images as moving objects. The detailed information for generating those video set for discovery is given in Appendix \ref{video_dataset}. The closed-form governing equations for those nonlinear dynamical systems are uncovered after completing the four-step training procedure introduced above. The experiments are performed on a workstation with 4 Tesla V100 GPU cards.

\begin{figure}
	\centering
	\includegraphics[width=0.8\linewidth]{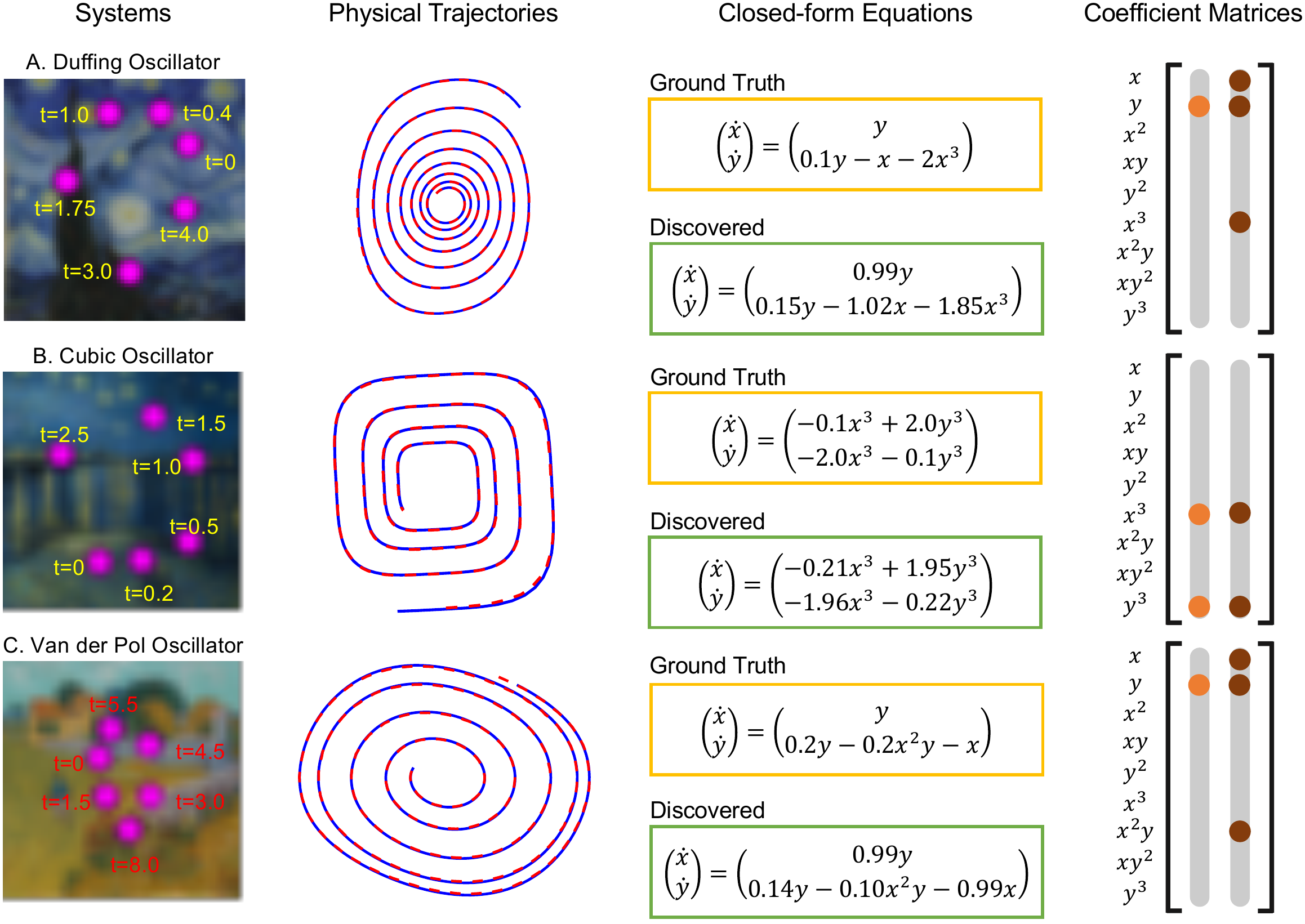}
	\caption{First order dynamical system discovery. The left shows several snapshots of the moving object in the videos. The ground truth and the learned physical trajectories are presented by solid and dash lines, respectively. The physical trajectories and closed-form equations are the results after scaling and variable transformation.}
	\label{fig:first_order_systems}
	\vspace{-8pt}
\end{figure}

\begin{figure}
	\centering
	\includegraphics[width=0.85\linewidth]{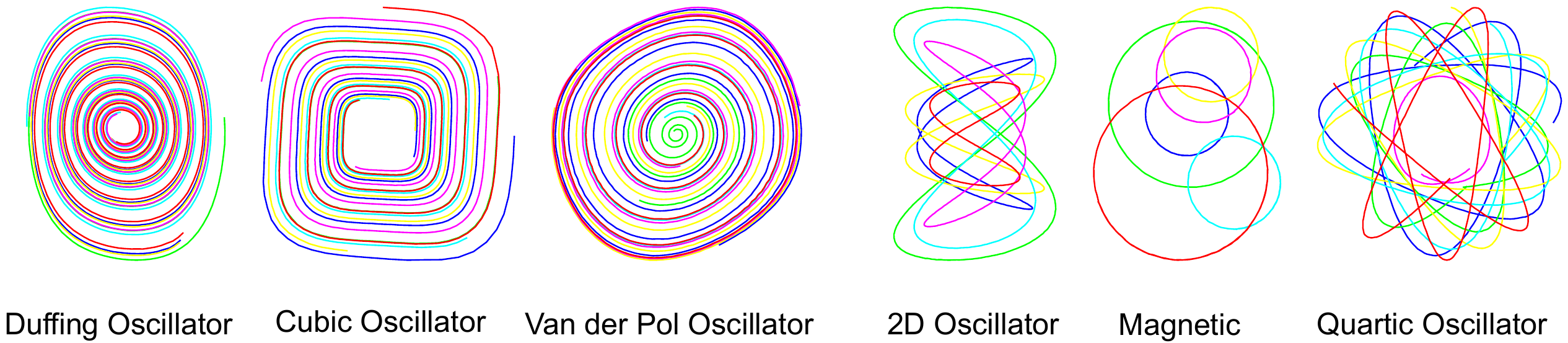}
	\caption{Unscaled physical trajectories of first and second order systems learned from network.}
	\label{fig:trajectories}
	\vspace{-8pt}
\end{figure}

\paragraph{First Order Systems}
The discovery result for the first order nonlinear dynamical systems is shown in Fig. \ref{fig:first_order_systems}. In our method, both physical trajectories and their governed equations are uncovered. For the first order systems, the physical model order is set to be 1 and the highest order of candidate polynomials is 3. It should be highlighted that, because of the Spatial-Physical Transformation which has translation factor for physical states, this method is not able to model the dynamical systems with a constant term. After pre-training, the autoencoder learns the spatial coordinate of moving object with reconstruction loss less than \(1.0\times10^{-5}\). The small reconstruction loss demonstrates the designed autoencoder is able to extract the latent space and reconstruct the video frames from latent space. In the total loss optimization, both physical derivative loss \(\mathcal{L}_{\dot{\mathbf{x}}_p}\), and forward and backward frames reconstruction loss \(\mathcal{L}_{int}\) converged along the update of candidate function coefficients \(\mathbf{\Xi}\). After training with the total loss, all these discovered equations show parsimony, where the sparse coefficients \(\mathbf{\Xi}\) have many close-to-zero terms. With sequential thresholding and refinement training, those nonlinear dynamics systems are fully discovered with both physical trajectories and the corresponding physical laws. Fig. \ref{fig:first_order_systems} shows the identified governing equations which are very close to the ground truth including the candidate terms and their coefficients. In addition, due to the scaling effect of the extracted physical states, suitable variable transformations are applied to the discovered equations and extracted physical trajectories. The detailed information of the discovery for all first order nonlinear systems are given in Appendix \ref{example_systems}. More physical trajectories extracted from network for the these systems are given in Fig. \ref{fig:trajectories}. 

\begin{figure}
	\centering
	\includegraphics[width=0.80\linewidth]{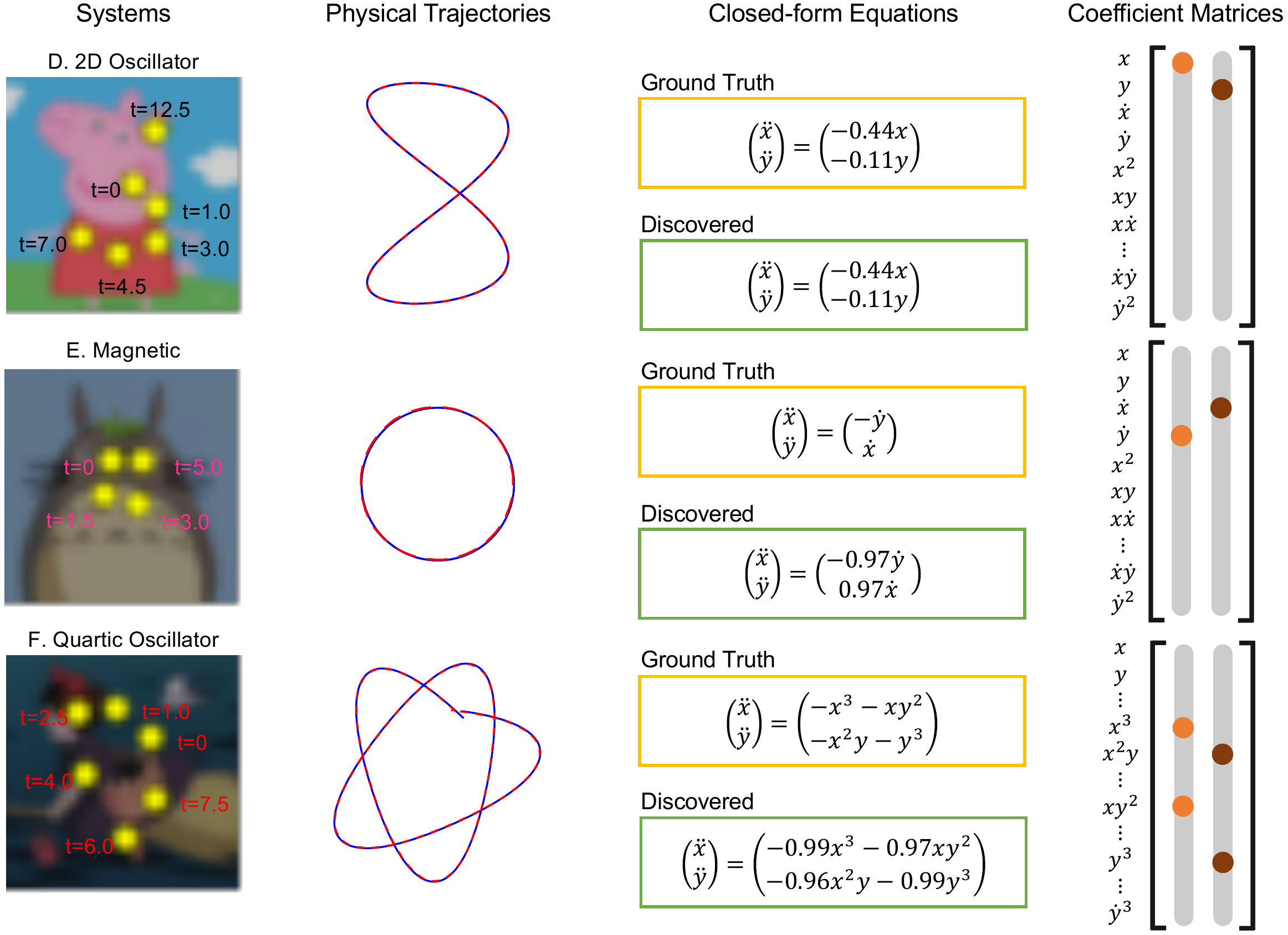}
	\caption{Second order system discovery. The results shown here have the same caption in Fig. \ref{fig:first_order_systems}.}
	\label{fig:second_order_systems}
	\vspace{-3pt}
\end{figure}

\paragraph{Second Order Systems}
The second order dynamics discovery shares the same framework and network with the first order systems in a state space form. In the modeling with state space, the autoencoder still learns the spatial coordinate of moving object which is then mapped into physical states by Spatial-Physical Transformation \(\mathcal{T}\). In the latent physical space, the extracted physical states \(\mathbf{x}_p=(x_p,y_p)\) are added with their first order derivative \(\mathbf{\dot{x}}_p=(\dot{x}_p,\dot{y}_p)\) using central difference, which results in four variables, \((x_p,y_p,\dot{x}_p,\dot{y}_p)\), in the physical modeling. This state space modeling enables the forward/backward propagation of the physical states by using Runge-Kutta time integration. Despite with four variables in the latent physical modeling, only physical state \((x_p,y_p)\) (its first order derivative \(\dot{x}_p,\dot{y}_p)\) excluded) is required for video frame reconstruction in the decoder. Besides, in order to simply the discovery, the derivative equations \(d\mathbf{x}_p/dt=\dot{\mathbf{x}}_p\) in state space are assumed to be known. The detailed implementation for the second order dynamical system discovery are given in Appendix \ref{example_systems}. Even if the state space modeling involves more candidate terms (including derivative terms) in the physical function library, the identified coefficient matrices \(\mathbf{\Xi}\) also show high sparsity. Fig. \ref{fig:second_order_systems} shows the results for the second order nonlinear dynamics discovered from videos. It can be seen that the governing equation coefficients for all systems are identified accurately. Fig. \ref{fig:trajectories} further shows the uncovered physical trajectories from different videos. 

\paragraph{Ablation Study.} \label{ablation_study}
In the proposed method, the autoencoder for physical states and spares regression for explicit physical law are trained in a joint optimization. The imposed physical law provides constraint to the autoencoder for physical state extraction by \(\mathcal{L}_{\dot{\mathbf{x}}_p}\) and \(\mathcal{L}_{int}\). Here, we designed an ablation study which decouples the joint optimization; namely, a two-step discovery is studied, where the extraction of physical states \(\mathbf{x}_p\) and sparse regression for modeling dynamics \(\frac{d}{dt}\mathbf{x}_p(t)=\bm{f}(\mathbf{x}_p(t))\) are conducted sequentially. This is used to validate the necessity of joint optimization and the importance of physical law constraint on the autoencoder. The two-step discovery keeps the learning of spatial coordinate of moving object \(\mathbf{x}_s\) from pre-training in the complete method. Since the physical coordinate is unknown (e.g., the coordinate origin), the spatial coordinates are converted into physical states \(\mathbf{x}_p\) with given approximated transformation factors \(\tilde{s}\) and \(\tilde{t}\). The governing equation is then identified from these physical states by STRidge \cite{rudy2017data}, a sparse regression method. The detailed procedure of the two-step discovery is given in Appendix \ref{ablations}. We take Duffing Oscillator and 2D Oscillator as examples with the discovery summary result listed in Table. \ref{ablation_studies}. The false positive discovery result shows that, without physical constraint, the spatial coordinate from pure autoencoder is not able to uncover the correct governing equations.

We also performed two other ablation cases: (1) remove the Spatial-Physical Transformation in autoencoder, and (2) remove the forward and backward frame reconstructions loss \(\mathcal{L}_{int}\). We take Cubic Oscillator and Magnetic systems as examples for the ablation without Spatial-Physical Transformation. When the network is optimized with total loss function, \(\mathcal{L}_{\dot{\mathbf{x}}_p}\) and \(\mathcal{L}_{int}\) blow up and do not converge. This is because adding physical law constraint to spatial coordinate directly will not help the autoencoder to extract the physical states but lead to loss divergence. For the ablation test without \(\mathcal{L}_{int}\), Van der Pol Oscillator and Quartic Oscillator systems are taken as examples. After training the network, it is found that \(\mathcal{L}_{int}\) remains large indicating that the time evolution of the dynamics is not well satisfied. The discovery results of the ablation networks are shown in Table \ref{ablation_studies}.

\begin{table}
\caption{Ablation study result.}
\label{ablation_studies}
\centering
{\small
\begin{tabular}{lll|ll|ll} 
\toprule
& \multicolumn{2}{c|}{Two-step Discovery}  & \multicolumn{2}{c|}{Remove \(\mathcal{T}\) and \(\tilde{\mathcal{T}}\)} & \multicolumn{2}{c}{Remove \(\mathcal{L}_{int}\)}               \\ 
\cmidrule{2-7}
Results & Duffing & 2D Oscillator & Cubic & Magnetic & Van der Pol  & Quartic  \\ 
\midrule
\# of TPT$^\text{a}$ for \(x\)-eq  & 1 [1]$^\text{c}$ & 0 [1] & NA & NA & 1 [1] &  2 [2] \\
\midrule
\# of TPT for \(y\)-eq  & 2 [3] & 1 [1] & NA & NA & 2 [3] & 1 [2]\\
\midrule
\# of FPT$^\text{b}$ for \(x\)-eq  & 6 & 5 & NA & NA & 1 & 2 \\
\midrule
\# of FPT for \(y\)-eq  & 5 & 0 & NA & NA & 1 & 4  \\
\bottomrule
\end{tabular}
}
\begin{tablenotes}
\item{\small Note: a. TPT denotes ``true positive terms''. b. FPT represents ``false positive terms''. c. This number in the square brackets depicts the reference (e.g., the correct number of terms in the ground truth equation).}
\end{tablenotes}
\vspace{-8pt}
\end{table}

\paragraph{Baseline Comparison.} \label{baseline_comparison}

We found that the literature remains scant in the present topic except for the closely-related method proposed in \cite{champion2019data}, which dealt with coordinate and governing equation discovery from high-dimensional data. We take this method as a baseline model. Since such a method requires the temporal derivative of input, the video is converted into gray scale for derivative calculation. 
The result (see Appendix \ref{baselines}) shows that the baseline model failed to both extract the physical trajectory of moving object and discover the correct equations. 
Appendix \ref{baselines} gives the detailed discovery results for Duffing Oscillator by using this method compared with ours.

\section{Discussion and Conclusion} \label{discussion_future}

We present an end-to-end unsupervised deep learning scheme to simultaneously discover the physical states of moving object and the underlying governing equations (or called physical law) from raw videos. Our approach takes the advantage of the power of neural network by using an autoencoder architecture to learn the spatial/pixel coordinate of moving object. A learnable Spatial-Physical Transformation maps the spatial/pixel coordinates into physical states where the dynamics may be explicitly expressed. The extracted physical states are described by a set of candidate functions with their learnable coefficients, which not only enables the physical law to be embedded into the network, but also provides a physical constraint to the autoencoder. This work addresses key limitations of prior approaches for physical law discovery from videos. We have successfully demonstrate the proposed method on a variety of nonlinear dynamical systems.

Our work also suggests several future research directions. First of all, real scenes of interest do not follow the assumption that both the moving object and background are constant (even though this assumption is widely used). Adapting our proposed paradigm to varying and non-stationary scene background would require some update for the autoencoder for learning the spatial coordinate of moving object in non-constant scene. The assumption of constant object and background also brings out a future work on testing the robustness of the proposed on noisy videos. In addition, since the autoencoder employed in this paper has the capacity of multiple object location, the proposed method has the potential to uncover the explicit dynamics presented by multiple objects. Besides, the videos studied in this paper has no warp and the coordinate relationship between spatial coordinate of moving object and physical states is very simple without considering rotation. However, the imposition of physical law constraint would allow for the extraction of more complex physical states (e.g., with distortion and rotation) and their governed equations. These issues will be studied in the future, on the basis of using real-recorded videos.

\section{Broader Impact} \label{broader_mpact}

The proposed method provides a way of explicit closed-form governing equation discovery for nonlinear dynamics directly from raw videos. 
Overall, our work sets forward a new idea of discovery of parsimonious interpretable models in a flexible and accessible sensing environment where only videos are available. Such an approach can be potentially used in many scientific disciplines, where imaging/video observations are available and meanwhile analytical models are in great need.



\renewcommand{\bibsection}{}
\section*{References}
\vspace{1pt}
\small
\bibliographystyle{unsrt}
\bibliography{references}

\clearpage

\appendix

\renewcommand\thesection{\Alph{section}}
\renewcommand\thesubsection{\thesection.\arabic{subsection}}
\renewcommand\thesubsubsection{\thesubsection.\arabic{subsubsection}}

\setcounter{figure}{0} 
\setcounter{table}{0} 
\setcounter{equation}{0}
\renewcommand{\thefigure}{A.\arabic{figure}}
\renewcommand{\thetable}{A.\arabic{table}}
\renewcommand{\theequation}{A.\arabic{equation}}

\section*{Appendix} \label{appendix}

This appendix provides a detailed description of the dataset generation, the proposed method which includes the designed network architecture, loss function definition, choice of hyperparameters and training procedure, the studied examples, ablation study and baseline comparison.

\section{Video Dataset} \label{video_dataset}
The videos studied here are synthesized to imitate the real physical videos in MATLAB. At first, the dynamic system is given and its trajectory is simulated with function ode113. With horizontal and vertical coordinates, each point in trajectory is plotted as a circle marker in 2D coordinate system, generating a color image with object only and with size of \(256\times256\). The object image is downsampled into \(64\times64\) later. Then a color image with size of \(64\times64\) is randomly selected to simulate the background in real videos. In the background image, the pixels at the location of moving object are removed and replaced by the downsampled object. The generated videos are colorful videos with RGB channels.

\section{Closed-form Governing Equation Discovery from Videos} \label{Closed_form _overning_equation_discovery_from_videos}
\subsection{Network Architecture} \label{network_architecture_appendix}

We present a novel end-to-end unsupervised deep learning paradigm with sparse regression to simultaneously extract the physical states and identify their governed parsimonious closed-form equation for nonlinear dynamics presented by moving object in videos. The high-level view of the designed network shown in \ref{fig:network_architecture} (A) involves the autoencoder for physical states extraction from video and the imposition of physical law constraint. The high-level view network consists of four basic components which are Encoder \(\varphi\), Coordinate-Consistent Decoder \(\psi\), spatial to physical coordinate transformation \(\mathcal{T}\) (and physical to spatial coordinate transformation \(\tilde{\mathcal{T}}\)), and sparse regression on physical states.

For the Encoder, the video frames are fed into a U-Net which constitutes several convolution layers and max pooling layers to produce unnormalized masks which have the same size as the input frame and represent the location of moving object. The unnormalized object mask is stacked with constant mask for background and then passed through a Softmax to generate masks each of which is assigned to a pixel in the video frame. The pixel-wise multiplication of the input video frame and the masks are flatten and fed into a localization network which is made of 3 fully-connected layers (200 units in hidden layers and 2 units in the output layer). The output 2 elements from the dense network is then passed through a saturating non-linear activation function \(H/2\cdot\tanh{(\cdot)}+H/2\) to produce the pixel coordinate of the moving object at two dimensions in the video frame with size of \(H \times H\). Here, the volume of U-Net can be fine-tuned by changing the number of layers and number of channels in each layer.

The decoder takes the spatial coordinate of moving object \(\mathbf{x}_s\) as input given by the Physical-Spatial Transformation of physical state and outputs a reconstructed image. Instead of using conventional decoders (like convolutional decoder), the coordinate consistent decoder which is imposed by a correct relationship between latent spatial coordiante and pixel-coordiante correspondence is employed to reconstruct the video frame. The Coordinate-Consistent Decoder is based on the spatial transformer (ST) proposed in \cite{jaques2020physics}. If the input spatial coordinate to decoder is \(\mathbf{x}_s=(x_s,y_s)\), the parameters \(\omega\) of ST is to place the center of the writing attention window for moving object at this point. The Coordinate-Consistent Decoder proposed in \cite{jaques2020physics} uses ST with inverse transformation parameters to represent the translation, scale or angle of the writing attention window. For a general affine transformation with translation \((t_x,t_y)\), angle \(\vartheta\) and scale \(s\), the source image coordinates can be modified according to:
\begin{equation}
\label{source_coordinates_modification}
\begin{pmatrix}x_s^o\\y_s^o\\1\end{pmatrix}=
\begin{pmatrix}s\cdot\cos\vartheta& \ s\cdot\sin\vartheta& \ t_x& \\
-s\cdot\sin\vartheta& \ s\cdot\cos\vartheta& \ t_y&\\
0& \ 0& \ 1&\\ \end{pmatrix}
\begin{pmatrix}x_s^s\\y_s^s\\1\end{pmatrix}
\end{equation}
where \((x_s^o,y_s^o)\) and \((x_s^s,y_s^s)\) are the output and source image coordinates. This transformation can be obtained with a ST by inverting Equation \ref{source_coordinates_modification}:
\begin{equation}
\label{st_inverse_transformation}
\begin{pmatrix}x_s^s\\y_s^s\\1\end{pmatrix}=
\frac{1}{s}
\begin{pmatrix}\cos\vartheta& \ -\sin\vartheta& \ -t_x\cos\theta+t_y\sin\vartheta& \\
\sin\vartheta& \ \cos\vartheta& \ -t_x\sin\vartheta-t_y\cos\vartheta& \\
0& \ 0& \ s&\\ \end{pmatrix}
\begin{pmatrix}x_s^o\\y_s^o\\1\end{pmatrix}
\end{equation}
Therefore, in the Coordinate-Consistent Decoder, the image can be reconstructed by using the ST with parameters \(\omega\) given in Equation \ref{st_inverse_transformation}. Each video frame can be reconstructed by the combination of moving object and background, both of which can be learned by content and mask. The moving object is represented by a learnable content \(\mathbf{c}\in[0,1]^{H \times H \times C}\) and mask tensor \(\mathbf{m}\in\mathbb{R}^{H \times H \times 1}\). The background is also represented by a content \(\mathbf{c}^b\in[0,1]^{H \times H \times C}\), but a mask \(\mathbf{m}^b\) which is a tensor with ones. The content and mask of the moving object are transformed with ST as \([\hat{\mathbf{c}},\hat{\mathbf{m}}]=\mathrm{ST}([\mathbf{c},\mathbf{m}],\omega)\). The resulting logit mask for moving object is stacked with the constant background mask \(\mathbf{m}^b\) and passed into a Softmax, \([\Tilde{\mathbf{m}},\Tilde{\mathbf{m}}^b]=\mathrm{softmax}(\mathbf{m},\mathbf{m}^b)\). Then the output image can be represented as the summation of the multiplications of contents and masks for moving object and background:
\begin{equation}
\label{decoder_output}
\Tilde{\mathbf{I}} = \tilde{\mathbf{m}}^b\odot\mathbf{c}^b + \tilde{\mathbf{m}}\odot\mathbf{\hat{\mathbf{c}}} 
\end{equation}
In the Coordinate-Consistent Decoder, both the content and mask are learned by independent networks. These fully-connected networks have a constant array of 1s as input and 1 hidden layer with 200 nodes for the object and 500 nodes for the background. The outputs of these networks are activated with \(\tanh\) and reshaped into 2D content and mask.

The learned spatial coordinate of moving object is then passed through a spatial to physical coordinate transformer to produce the expected physical states which may be responsible to parsimonious closed form governing equations. If we assume the dynamic system is built in the 2D plane image space, the transformation between spatial and physical coordinate can be implemented by standard 2D Cartesian coordinate transformation. In coordinate transformation, the position of moving object is kept fixed while the coordinate system is transformed relative to the object. The general coordinate transformation involves translation, scaling and rotation. If the transformation has translation vector \(\tilde{\mathbf{t}}=(\tilde{t}_x,\tilde{t}_y)^T\), scaling factor \(\tilde{s}\) and rotation angle \(\tilde{\vartheta}\), the physical coordinate can be expressed from spatial coordinate as
\begin{equation}
\label{coordinate_transformation}
\mathbf{x}^T_p=\mathcal{T}(\mathbf{x}^T_s)=\tilde{s}\big[(\mathbf{Q}(\tilde{\vartheta})\mathbf{x}^T_s-\tilde{\mathbf{t}})\big]
\end{equation}
where \(\mathbf{Q}(\tilde{\vartheta})\) is the transformation matrix induced by rotation angle \(\vartheta\), which is
\begin{equation}
\label{transformation_matrix_rotation}
\mathbf{Q}(\tilde{\vartheta})=
\begin{pmatrix}\cos\tilde{\vartheta} & \sin\tilde{\vartheta} \\ -\sin\tilde{\vartheta} & \cos\tilde{\vartheta} \\
\end{pmatrix}
\end{equation}
Then the transformation can be expanded as
\begin{equation}
\label{coordinate_transformation_expansion}
\begin{pmatrix}x_p\\y_p\end{pmatrix}=
\mathcal{T}(\mathbf{x}^T_s)=\tilde{s}
\begin{pmatrix}\cos\tilde{\vartheta}\cdot{x_s}+\sin\tilde{\vartheta}\cdot{y_s}-\tilde{t}_x \\ -\sin\tilde{\vartheta}\cdot{x_s}+\cos\tilde{\vartheta}\cdot{y_s}-\tilde{t}_y \\
\end{pmatrix}
\end{equation}

In the dataset studied here, the real physical coordinate system is parallel to  the spatial coordinate and the coordinate transformation has no rotation, the spatial to physical transformation is degraded into 
\begin{equation}
\label{coordinate_transformation_degrade_expansion}
\begin{pmatrix}x_p\\y_p\end{pmatrix}=
\mathcal{T}(\mathbf{x}^T_s)=\tilde{s}
\begin{pmatrix}{x_s}-\tilde{t}_x \\ {y_s}-\tilde{t}_y \\
\end{pmatrix}
\end{equation}
 Likewise, the physical to spatial transformation can be simplified as
\begin{equation}
\label{physical_to_spatial_degrade}
\begin{pmatrix}x_s\\y_s\end{pmatrix}=
\tilde{\mathcal{T}}(\mathbf{x}^T_p)=
\begin{pmatrix}1/\tilde{s}\cdot x_p+\tilde{t}_x \\ 1/\tilde{s}\cdot y_p+\tilde{t}_y\\
\end{pmatrix}
\end{equation}
Here \(\mathcal{T}\) and \(\tilde{\mathcal{T}}\) share the same transformation factors \(\tilde{s}\), \(\tilde{\mathbf{t}}\) and \(\tilde{\vartheta}\). Those factors are trainable variables and will be optimized in the network training.

\subsection{Loss Functions} \label{loss_function}
The loss function used in training is a weighted sum of four terms: autoencoder reconstruction loss \(\mathcal{L}_{recon}\), physical state derivative loss \(\mathcal{L}_{d\mathbf{x}/dt}\), forward and backward video frame reconstruction loss from integral physical state \(\mathcal{L}_{int}\), and candidate function coefficients regularization \(\mathcal{L}_{reg}\). For one single input video \(I\) with \(N\) frames, each loss is explicitly defined as follows:
\begin{equation}
\label{autoencoder_loss_term}
\mathcal{L}_{recon} = \frac{1}{N}\sum_{k=1}^{N}
\left\lVert{I}_k-\psi(\tilde{\mathcal{T}}(\mathcal{T}(\varphi(I_k))))\right\rVert_2^2
\end{equation}
\begin{equation}
\label{derivative_physical_states_loss_term}
\mathcal{L}_{\dot{\mathbf{x}}_p} = \frac{1}{N}\sum_{k=1}^{N}
\left\lVert\nabla_t(\mathcal{T}(\varphi(I_k)) - \mathbf{\Theta}((\mathcal{T}(\varphi(I_k)))^T)\mathbf{\Xi}\right\rVert_2^2
\end{equation}
\begin{equation}
\label{video_reconstruction_integral_physical_state}
\mathcal{L}_{int} = \frac{1}{N\times(2q+1)} \sum_{k=1}^{N}\sum_{i=-q}^{q}
\left\lVert{I}_{k+i}-\psi(\tilde{\mathcal{T}}(\mathcal{T}(\varphi(I_k)+\int_{t_k}^{t_{k+i}}\mathbf{\Theta}(\mathbf{x}(\tau))\mathbf{\Xi}d\tau))\right\rVert_2^2
\end{equation}
\begin{equation}
\label{regularization_loss_term}
\mathcal{L}_{reg} = \frac{1}{2n}
\lVert{\mathbf{\Xi}\rVert}_{0.5}
\end{equation}
The total loss function is
\begin{equation}
\label{total_loss_terms}
\mathcal{L}_{total} = \mathcal{L}_{recon} + \mathcal{L}_{\dot{\mathbf{x}}_p} + \lambda_2\mathcal{L}_{int} + \lambda_3\mathcal{L}_{reg}
\end{equation}
\(\mathcal{L}_{recon}\) ensures that the autoencoder can accurately reconstruct the video frame from the physical states of the moving object. \(\mathcal{L}_{\dot{\mathbf{x}}_p}\) and \(\mathcal{L}_{int}\) ensure that the discovered physical model captures the dynamics of the physical states extracted from autoencoder incorporated with spatial to physical and its inverse transformation. Here, \(q\) denotes the time-steps forward/backward for the integral of physical states. \(\mathcal{L}_{reg}\) promotes the sparsity of dynamic model.

\subsection{Second order Dynamic System Discovery} \label{second_order_dynamics_discovery}
The proposed method is also applicable to the discovery for second order dynamics by transferring the systems into state-space model. Second order systems discovery shares the same network with the first order but slight modification in the physical constraint and equation identification. In the network, Encoder still condenses the video frame into spatial coordinate of moving object \((x_s, y_s)\), which is followed by Spatial-Physical Transformation to get physical coordinate \((x_p, y_p)\). If we define the variables \(x_1=x_p\) and \(y_1=y_p\) and calculate the first order derivatives of \(x_1\) and \(y_1\) as \(\dot{x}_1=x_2\) and \(\dot{y}_1=y_2\). Then the second order equation group is transferred into 4 first order equations as
\begin{equation}
\label{state_space}
\begin{pmatrix}\dot{x}_1\\\dot{x}_2\\\dot{y}_1\\\dot{y}_2\end{pmatrix}=
\begin{pmatrix}{x}_2\\f(x_1,x_2,y_1,y_2)\\y_2\\g(x_1,x_2,y_1,y_2)\end{pmatrix}
\end{equation}
where \(f(x_1,x_2,y_1,y_2)\) and \(g(x_1,x_2,y_1,y_2)\) are the equations needed to be discovered. Noteworthy, the state-space model transfer also make the Runge-Kutta method applicable to the integral of physical states. Although both physical coordinate and its derivative are obtained in latent space, only physical coordinate is feed into decoder for video reconstruction. In addition, in order to simplify the discovery, the equation \(\dot{x}_1=x_2\) and \(\dot{y}_1=y_2\) are given in \ref{state_space} for the discovery.

\subsection{Choice of Hyperparameters} \label{choice_hyperparameters}
The training procedure described above requires the choice of several hyperparameters: chiefly the order of discovered ODEs, candidate function polynomial order for sparse regression, initial values for coordinate transformation, and the loss weight penalties \(\lambda_1\), \(\lambda_2\), \(\lambda_3\). The choice of these parameters greatly impacts the success of governing equation discovery. Here we outline some guidelines for choosing these parameters.

The most important hyperparameter choice is the order of discovered ODEs \(d\), as this impacts the interpretation of the extracted physical states and the associated dynamical model. In this paper we built a general method to discover both first and second dynamical systems and discussed the examples of these two types of discovery from videos. Nevertheless, the order may not be obvious in real system. Here we suggest using the sparsity of discovered candidate function coefficients to determine the order \(d\). Because choosing \(d\) that is not correct to the real system may still result in a valid model for the dynamics; however, obtaining a sparse model may be more difficult. One may want to train the models with order \(d=1\) and \(d=2\). The order leading to much more obvious sparsity in the coefficient matrix will be the determined order \(d\). 

Since the equations discovered here are identified by sparse regression, this method requires a choice of library functions to interpret the extracted physical states. All the examples shown here use polynomial library terms. But for general systems, the best library functions to use may be unknown, and choosing the wrong library functions can leads to the failure discovery. A recommended practice is to start with polynomial models, as many common physical models are represented by polynomials. In addition, the order of polynomial library is still a hyperparameter to tune. On one hand, we prefer to make the discovery more general by using more diverse polynomial candidate terms. On the other hand, balancing the increasing theoretical and computational complexity is crucial for applications. Although the higher order of the polynomial is, the more likely the exact terms will be uncovered from the video. Nevertheless, too large scale candidate library may obscure the sparsity of physical model and leads to increasing difficulty on simplest model discovery. As simple library function is typically easy to train, one may want to start by using lower polynomial order like 2. Once the sparsity has been found, the model can be trained by fine-tuning other hyperparameters while keeping the library functions.

The choice of initial values of spatial-physical coordiante transformation also affects the discovery. After pre-training, Encoder captures the spatial coordinate of moving object which is then converted into physical coordinate. With other loss terms being added, the Encoder and Spatial-Physical Transformer are contrained by the physical law for physical states extraction. If the initial values for spatial-physical coordiante transformation have huge discrepancy to the ground truth, physical law will not provide proper constraint on the physical states. A recommended way to initialize Spatial-Physical Transformation is to determine the transformation factors based on the spatial coordiante of moving object after pre-training. Take the studied videos in this paper for example, the initialized translation factor enables the origin of physical coordinate system at the centre of spatial coordiante represented trajectories. For the scaling factor, since the discovered physical trajectories and governing equations have a scaled relationship with the ground truth, the choice of scaling factor may not be a big issue. Because the physical coordinate is typically smaller than pixel coordinate, one may want to try the scaling factor with multiplier of 10.

The choice of loss weight penalties \(\lambda_1\), \(\lambda_2\), \(\lambda_3\) also has a significant impact on the success of the governing equation discovery. The loss weight \(\lambda_2\) can be determined at first. If the governing equation is discovered, the reconstructed video frames from the forward and backward integral physical states should have the same error as the autoencoder reconstruction loss, the first term in total loss function, so the loss weight \(\lambda_2\) is chosen as 1.0. Then loss weight \(\lambda_1\) is chosen to be three orders of magnitude less than 1.0, around \(10^{-3}\). Although the derivative of physical states loss promotes the physical constraint to the physical states extraction and the governing equation discovery, \(\lambda_1\) cannot be too large. Because too large \(\lambda_1\) will destroy the autoencoder pre-trained for spatial coordiante of moving object. Furthermore, too large \(\lambda_1\) encourages shrinking of the magnitude of extracted physical states. The third parameters \(\lambda_3\) determines the strength of the regularization of the sparse regression model coefficients \(\mathbf{\Xi}\) and thus affects the sparsity of the resulting models. If this value is too large, the model will be too simple and achieve poor prediction; if it is too small, the modes will be nonsparse and prone to overfitting. The coefficients regularization loss weight requires the most tuning and should typically be chosen by trying a range of values and assessing the level of sparsity in the discovered equation. 

\subsection{Training Procedure} \label{training_procedure}
There are four steps in the network training for closed-form governing equation discovery: pre-training, total loss optimization, sequential thresholding and refinement. In the network, the trainable variables include the weights in Encoder \(W_\varphi\), weights in decoder \(W_\psi\), Spatial-Physical Transformation factors \({T}\), and physical library function coefficient \(\mathbf{\Xi}\). In the pre-training, the network is trained by only optimizing the autoencoder reconstruction loss \(\mathcal{L}_{recon}\) ,and only \(W_\varphi\) and \(W_\psi\) being updated. The training ends up with a small value for autoencoder reconstruction loss. The pre-trained model is then loaded for the total loss training in which all trainable variables are optimized. After total loss training, both \(\mathcal{L}_{\dot{\mathbf{x}}_p}\) and \(\mathcal{L}_{int}\) are reduced to small value and the candidate function coefficients show high sparsity. Later, the total trained model is loaded for the following sequential thresholding. In sequential thresholding, the network is trained along thresholding the candidate function terms whose coefficients are smaller than the given threshold. Finally, the sequential thresholding model is loaded for the final refinement. In the refinement, the network is trained by only updating the coefficients of the candidate function terms left from sequential thresholding. It should be noted that, in the sequential thresholding, once the function terms are filtered out for their smaller coefficient than threshold, the coefficients for those terms are set as 0 and will not be trained in the following epochs. After refinement training, the dynamic model is discovered from the candidate function coefficients in sparse regression. 

\section{Code} \label{code}
We use the Python API for TensorFlow to implement and train our network. Our code is publicly available at github.com.

\section{Example Systems} \label{example_systems}
We demonstrate the efficacy of our methodology on discovering various dynamical systems from videos. In particular, we discovered the first order dynamical systems including \textit{Duffing Oscillator}, \textit{Cubic Oscillator} and \textit{Van der Pol Oscillator}, and second order dynamical systems including \textit{2D Oscillator}, \textit{Magnetic} and \textit{Quartic Oscillator}. Different systems are trained in the same network but with different hyperparameters. 

\subsection{Duffing Oscillator} \label{duffing_oscillator}
The governing equation of Duffing Oscillator is
\begin{equation}
\label{duffing_oscillator_gt}
\begin{pmatrix}\dot{x}\\\dot{y}\end{pmatrix}=
\begin{pmatrix}y\\-p_1y-p_2x-p_3x^3\end{pmatrix}
\end{equation}
with \(p_1=0.1\), \(p_2=1\), and \(p_3=2\). To generate the video set, we simulate Duffing Oscillator from random initial conditions with \(x(0),y(0)\in[-0.8,0.8]\). In order to increase the diversity of training dataset, the trajectories with initial conditions \(|x(0)|<0.75\) and \(|y(0)|<0.5\) are removed with 64 trajectories left in total. Hyperparameters used for training Duffing Oscillator model are shown in Table \ref{hyperparameter_duffing}. First of all, the network was trained with 100 epochs for pre-training with only parameters of encoder and decoder being updated. After pre-training, the autoencoder reconstruction loss is very small, around \(5.4\times10^{-6}\), which means the designed network (especially the volumes of U-Net, and the networks for contents and background learning) has the capacity to extract the latent space and reconstruct the video frames from latent space. The forward and backward frame reconstruction loss is \(3.4\times10^{-4}\) and physical loss 0.16. It means, with initial candidate function coefficients (0.1), the forward and backward reconstructed video frames are very close to the ground truth, which not only will make the following training easy to converge, but also verifies the better learning of video background in Coordinate-Consistent Decoder. Then the network was kept training with total loss for 500 epochs. In the total loss training, network decreases the physical loss and reduces the error of forward and backward reconstructed frames at the same time by updating sparse regression coefficients. During the total loss training, the autoencoder reconstruction loss is very small, less than \(1.0\times10^{-5}\) along the whole training. Keeping the autoencoder  reconstruction loss very small is one key to succeed the physical law discovery. Because it means the spatial coordiante of moving object extraction is kept after adding physical loss constraint. Besides, during the training, both physical loss and forward and backward frame reconstruction loss are reduced, which means the model is optimized in the same correct direction. After total loss training, the autoencoder reconstruction loss is \(4.1\times10^{-6}\), forward and backward frame reconstruction loss \(5.6\times10^{-6}\) and physical states derivative loss \(0.0036\). For sparse regression coefficients, only 4 expected terms have coefficient larger than 0.1. The trained total loss model is then loaded for the following sequential thresholding training. In sequential thresholding, the threshold value is 0.1 and sparse regression coefficients are thresholded every 100 epochs. After 300 epochs, the redundant terms are filtered out and only 4 terms kept. Finally, the network was kept training for 200 epochs with the left 4 candidate terms. The discovered closed-form equation for cubic system is
\begin{equation}
\label{duffing_discovery_discovery}
\begin{pmatrix}\dot{x}\\\dot{y}\end{pmatrix}=
\begin{pmatrix}0.99y\\-0.15y-1.02x-3.73x^3\end{pmatrix}
\end{equation}
Then a suitable variable transformation is applied to the discovered equation to eliminate the effect of scaling, which is similar to the transformation in \cite{champion2019data}. After variable transformation with scaling factor 1.42, the discovered equation can be rewritten as
\begin{equation}
\label{duffing_transformation}
\begin{pmatrix}\dot{x}\\\dot{y}\end{pmatrix}=
\begin{pmatrix}0.99y\\-0.15y-1.02x-1.85x^3\end{pmatrix}
\end{equation}
Likewise, the extracted physical trajectory also needs scaling to make it identical to the ground truth. The comparison of ground truth and scaled extracted physical trajectories is given in Fig \ref{fig:first_order_trajectories}.

\begin{table}
  \caption{Hyperparameter values for the Duffing Oscillator}
  \label{hyperparameter_duffing}
  \centering
  \begin{tabular}{lll}
    \toprule
    Parameter  &  Value   \\
    \midrule
    video number  &  64   \\
    video length  &  800  \\
    time step     &  0.05 \\
    \(q\)         &  3 \\
    batch size    &  1    \\
    learning rate &   \(10^{-3}\) \\
    Dynamic model order & 1  \\
    Candidate function polynomial order  &  3   \\
    Physical states derivative loss \(\mathcal{L}_{\dot{\mathbf{x}}_p}\) weight \(\lambda_1\) &  \(1\times10^{-3}\)  \\
    Forward/backward frame reconstruction loss \(\mathcal{L}_{int}\) weight \(\lambda_2\) &  1.0  \\
    Candidate function coefficient regularization loss \(\mathcal{L}_{reg}\) weight \(\lambda_3\) &  \(5\times10^{-3}\)  \\
    \bottomrule
  \end{tabular}
\end{table}

\subsection{Cubic Oscillator} \label{cubic_oscillator} 
The governing equation of Cubic Oscillator is
\begin{equation}
\label{cubic_oscillator_gt}
\begin{pmatrix}\dot{x}\\\dot{y}\end{pmatrix}=
\begin{pmatrix}p_1x^3+p_2y^3\\p_3x^3+p_4y^3\end{pmatrix}
\end{equation}
with \(p_1=-0.1\), \(p_2=2\), \(p_3=-2\) and \(p_4=-0.1\). In the Cubic video set, the simulated Cubic Oscillators have initial conditions with \(x(0),y(0)\in[-1.2,1.2]\) and the trajectories with larger absolution value of \(x(0),y(0)\) less than 1.1 are removed with 64 trajectories left in total. Hyperparameters used for training Cubic Oscillator model are shown in Table \ref{hyperparameter_cubic}. After pre-training, the autoencoder reconstruction loss is \(5.5\times10^{-6}\), the forward and backward frame reconstruction loss \(1.8\times10^{-4}\), and  physical state derivative loss 0.15. Then the network was kept training with total loss for 300 epochs. After total loss training, the autoencoder reconstruction loss is \(3.7\times10^{-6}\), forward and backward frame reconstruction loss \(5.9\times10^{-6}\) and physical state derivative loss \(0.0051\). For sparse regression coefficients, there are 6 values larger than 0.1. The trained total loss model is then loaded for the following sequential thresholding training. In sequential thresholding, the threshold value is 0.1 and sparse regression coefficients are thresholded every 100 epochs. After 300 epochs, the redundant terms are filtered out and only 4 terms kept. Finally, the network was kept training for 200 epochs with the left 4 candidate terms. The discovered closed-form equation for Cubic Oscillator is
\begin{equation}
\label{cubic_oscillator_discovery}
\begin{pmatrix}\dot{x}\\\dot{y}\end{pmatrix}=
\begin{pmatrix}-0.33x^3+3.05y^3\\-3.07x^3-0.35y^3\end{pmatrix}
\end{equation}
Then the variable transformation is applied to the discovered equation to eliminate the effect of scaling. After transformation with scaling factor 1.28, the discovered equation can be rewritten as
\begin{equation}
\label{cubic_transformation}
\begin{pmatrix}\dot{x}\\\dot{y}\end{pmatrix}=
\begin{pmatrix}-0.21x^3+1.95y^3\\-1.96x^3-0.22y^3\end{pmatrix}
\end{equation}
Likewise, the extracted physical trajectory also needs scaling to be identical to the ground truth. The comparison of ground truth and scaled extracted physical trajectories is given in Fig \ref{fig:first_order_trajectories}.

\begin{table}
  \caption{Hyperparameter values for the Cubic Oscillator}
  \label{hyperparameter_cubic}
  \centering
  \begin{tabular}{lll}
    \toprule
    Parameter  &  Value   \\
    \midrule
    video number  &  64   \\
    video length  &  1001  \\
    time step     &  0.05 \\
    \(q\)         &  3 \\
    batch size    &  2    \\
    learning rate &   \(10^{-3}\) \\
    Dynamic model order & 1  \\
    Candidate function polynomial order  &  3   \\
    Physical states derivative loss \(\mathcal{L}_{\dot{\mathbf{x}}_p}\) weight \(\lambda_1\) &  \(1\times10^{-3}\)  \\
    Forward/backward frame reconstruction loss \(\mathcal{L}_{int}\) weight \(\lambda_2\) &  1.0  \\
    Candidate function coefficient regularization loss \(\mathcal{L}_{reg}\) weight \(\lambda_3\) &  \(2\times10^{-3}\)  \\
    \bottomrule
  \end{tabular}
\end{table}

\subsection{Van der Pol Oscillator} \label{vdp_oscillator}
\begin{figure}
	\centering
	\includegraphics[width=1.0\linewidth]{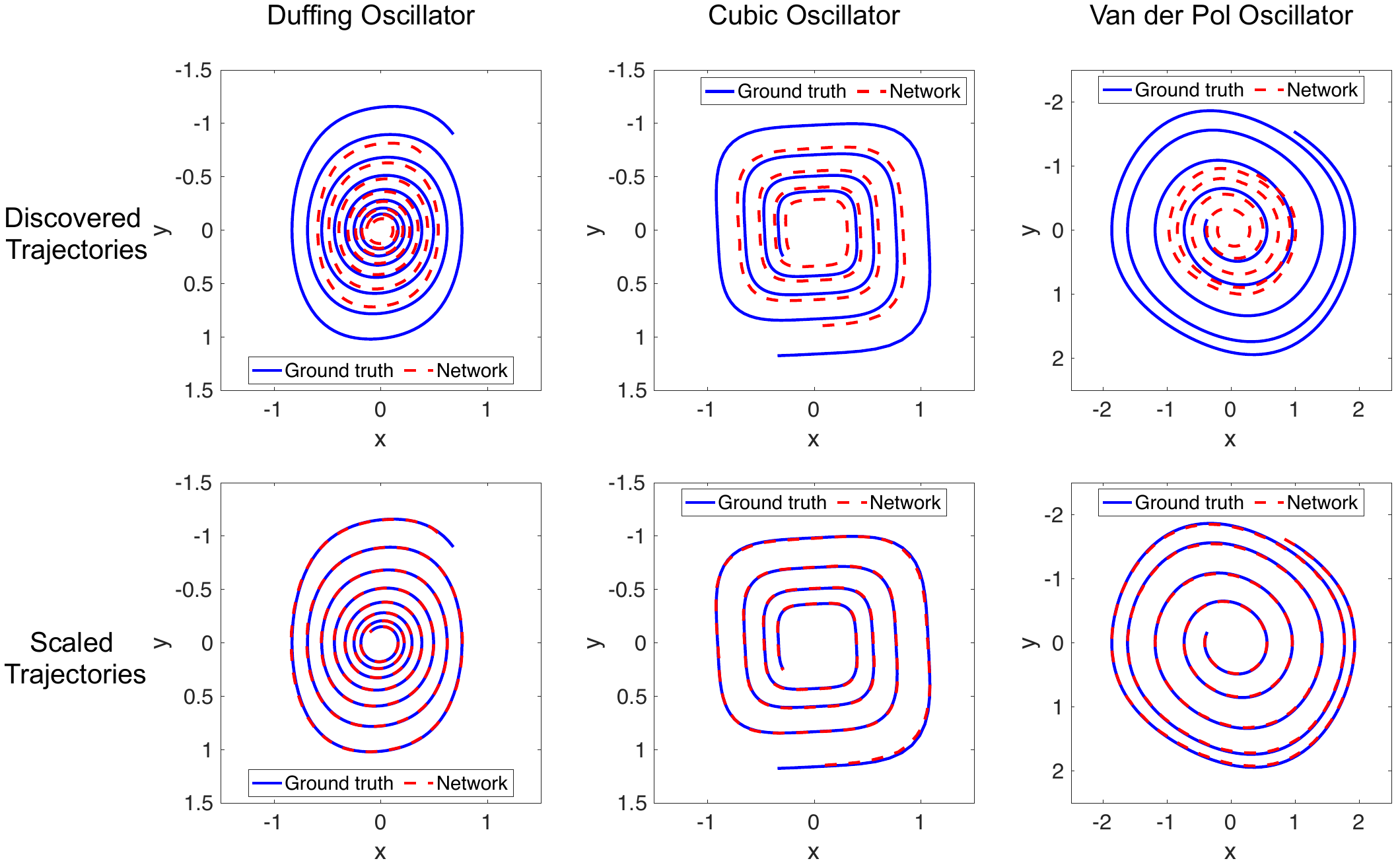}
	\caption{Discovered and scaled physical trajectories for first order systems extracted from network.}
	\label{fig:first_order_trajectories}
\end{figure}

The governing equation of Van der Pol Oscillator is
\begin{equation}
\label{vdp_oscillator_gt}
\begin{pmatrix}\dot{x}\\\dot{y}\end{pmatrix}=
\begin{pmatrix}y\\\mu(1-x^2)y-x\end{pmatrix}
\end{equation}
with \(\mu=0.2\). The video set is generated by simulating Van der Pol Oscillator from random initial conditions  \(x(0),y(0)\in[-0.5,0.5]\) with 64 trajectories in total. Hyperparameters used for training Van der Pol Oscillator model are shown in Table \ref{hyperparameter_vdp}. The network was trained with 300 epochs for pre-training. After pre-training, the autoencoder reconstruction loss, forward and backward frame reconstruction loss and physical state derivative loss are is \(7.5\times10^{-6}\), \(4.1\times10^{-4}\) and \(0.74\). Then the network was trained with total loss for 300 epochs. After total loss training, the autoencoder reconstruction loss is \(6.8\times10^{-6}\), forward and backward frame reconstruction loss \(9.1\times10^{-6}\) and physical state derivative loss \(0.0023\). For sparse regression coefficients, there are 4 values larger than 0.1, which means the coefficients for all redundant terms are less than 0.1. With threshold value 0.1, those redundant terms are filtered out in the first sequential thresholding. Finally, the network was kept training for 200 epochs with the left 4 candidate terms. The discovered closed-form equation for cubic system is
\begin{equation}
\label{vdp_discovery}
\begin{pmatrix}\dot{x}\\\dot{y}\end{pmatrix}=
\begin{pmatrix}0.99y\\-0.99x+0.14y-0.68x^2y\end{pmatrix}
\end{equation}
After variable transformation with scaling factor 1.92, the discovered Van der Pol system can be rewritten as
\begin{equation}
\label{vdp_transformation}
\begin{pmatrix}\dot{x}\\\dot{y}\end{pmatrix}=
\begin{pmatrix}0.99y\\-0.99x+0.14y-0.10x^2y\end{pmatrix}
\end{equation}
Likewise, the extracted physical trajectory is also scaled and the comparison of scaled extracted physical trajectory and ground truth is given in Fig \ref{fig:first_order_trajectories}.

\begin{table}
  \caption{Hyperparameter values for the Van der Pol Oscillator}
  \label{hyperparameter_vdp}
  \centering
  \begin{tabular}{lll}
    \toprule
    Parameter  &  Value   \\
    \midrule
    video number  &  64   \\
    video length  &  601  \\
    time step     &  0.05 \\
    \(q\)         &  3 \\
    batch size    &  2    \\
    learning rate &   \(10^{-3}\) \\
    Dynamic model order & 1  \\
    Candidate function polynomial order  &  3   \\
    Physical states derivative loss \(\mathcal{L}_{\dot{\mathbf{x}}_p}\) weight \(\lambda_1\) &  \(1\times10^{-3}\)  \\
    Forward/backward frame reconstruction loss \(\mathcal{L}_{int}\) weight \(\lambda_2\) &  1.0  \\
    Candidate function coefficient regularization loss \(\mathcal{L}_{reg}\) weight \(\lambda_3\) &  \(5\times10^{-3}\)  \\
    \bottomrule
  \end{tabular}
\end{table}

\subsection{2D Oscillator} \label{2d_oscillator}
The governing equation of 2D Oscillator is
\begin{equation}
\label{2d_oscillator_gt}
\begin{pmatrix}\ddot{x}\\\ddot{y}\end{pmatrix}=
\begin{pmatrix}-\cfrac{4}{9}x\\-\cfrac{1}{9}y\end{pmatrix}
\end{equation}
The video set for 2D Oscillator is generated from random initial conditions  \(x(0),y(0),\dot{x}(0),\dot{y}(0)\in[-1.0,1.0]\) with 128 trajectories in total. Hyperparameters used for training 2D Oscillator model are shown in Table \ref{hyperparameter_2d_oscillator}. The network was trained with 300 epochs for pre-training. After pre-training, the autoencoder reconstruction loss, forward and backward frame reconstruction loss and physical states derivative loss are is \(5.9\times10^{-6}\), \(7.2\times10^{-6}\) and \(0.98\). After total loss training, the autoencoder reconstruction loss is \(5.1\times10^{-6}\), forward and backward frame reconstruction loss \(5.1\times10^{-6}\) and physical state derivative loss \(0.0049\). For sparse regression coefficients, only 2 terms have coefficient larger than 0.1. With threshold value 0.1, all redundant terms are filtered out in the first sequential thresholding. Finally, the network was kept training for 200 epochs with the left 2 candidate terms. The discovered closed-form equation for cubic system is
\begin{equation}
\label{2d_oscillator_discovery}
\begin{pmatrix}\ddot{x}\\\ddot{y}\end{pmatrix}=
\begin{pmatrix}-0.44x\\-0.11y\end{pmatrix}
\end{equation}
From the extracted physical trajectory, a variable transformation with scaling factor 2.33 is still needed to make it identical to the ground truth. But the transformed governing equation is kept the same. The  comparison of scaled extracted and ground truth physical trajectory is given in Fig \ref{fig:second_order_trajectories}.

\begin{table}
  \caption{Hyperparameter values for the 2D Oscillator}
  \label{hyperparameter_2d_oscillator}
  \centering
  \begin{tabular}{lll}
    \toprule
    Parameter  &  Value   \\
    \midrule
    video number  &  128   \\
    video length  &  401  \\
    time step     &  0.05 \\
    \(q\)         &  3 \\
    batch size    &  2    \\
    learning rate &   \(10^{-3}\) \\
    Dynamic model order & 2  \\
    Candidate function polynomial order  &  2   \\
    Physical states derivative loss \(\mathcal{L}_{\dot{\mathbf{x}}_p}\) weight \(\lambda_1\) &  \(1\times10^{-3}\)  \\
    Forward/backward frame reconstruction loss \(\mathcal{L}_{int}\) weight \(\lambda_2\) &  1.0  \\
    Candidate function coefficient regularization loss \(\mathcal{L}_{reg}\) weight \(\lambda_3\) &  \(5\times10^{-3}\)  \\
    \bottomrule
  \end{tabular}
\end{table}

\subsection{Magnetic} \label{magnetic} 
The governing equation of Magnetic system is
\begin{equation}
\label{magnetic_gt}
\begin{pmatrix}\ddot{x}\\\ddot{y}\end{pmatrix}=
\begin{pmatrix}\dot{y}\\-\dot{x}\end{pmatrix}
\end{equation}
The video set for Magnetic is generated from random initial conditions  \(x(0),y(0),\dot{x}(0),\dot{y}(0)\in[-0.5,0.5]\) with 256 trajectories in total. Hyperparameters used for training Magnetic model are shown in Table \ref{hyperparameter_magnetic}. The network was trained with 300 epochs for pre-training. After pre-training, the autoencoder reconstruction loss, forward and backward frame reconstruction loss and physical state derivative loss are \(3.7\times10^{-6}\), \(5.6\times10^{-6}\) and \(0.44\). After 500 iterations of total loss training, the autoencoder reconstruction loss is \(3.3\times10^{-6}\), forward and backward frame reconstruction loss \(3.3\times10^{-6}\) and physical state derivative loss \(0.0029\). For sparse regression coefficients, there are 4 terms having coefficient larger than 0.1. The coefficients of redundant terms are small, about 0.13, and these redundant terms are filtered out after sequential thresholding. Finally, the network was kept training for 200 epochs with left 2 candidate terms. The discovered closed-form equation for Magnetic system is
\begin{equation}
\label{magnetic_discovery}
\begin{pmatrix}\ddot{x}\\\ddot{y}\end{pmatrix}=
\begin{pmatrix}0.97\dot{y}\\-0.97\dot{x}\end{pmatrix}
\end{equation}
Since the origin of magnetic system is not at \((0,0)\), The extracted Magnetic trajectory needs both scaling and translation factors to make it identical to the ground truth. For the given example, the extracted trajectory is scaled with factor 1.78, and then translated with \((1.30, 0.27)\). But the transformed governing equation is kept the same. The  comparison of scaled extracted and ground truth Magnetic trajectory is given in Fig \ref{fig:second_order_trajectories}.

\begin{table}
  \caption{Hyperparameter values for the Magnetic}
  \label{hyperparameter_magnetic}
  \centering
  \begin{tabular}{lll}
    \toprule
    Parameter  &  Value   \\
    \midrule
    video number  &  256   \\
    video length  &  201  \\
    time step     &  0.05 \\
    \(q\)         &  3 \\
    batch size    &  8    \\
    learning rate &   \(10^{-3}\) \\
    Dynamic model order & 2  \\
    Candidate function polynomial order  &  2   \\
    Physical states derivative loss \(\mathcal{L}_{\dot{\mathbf{x}}_p}\) weight \(\lambda_1\) &  \(1\times10^{-3}\)  \\
    Forward/backward frame reconstruction loss \(\mathcal{L}_{int}\) weight \(\lambda_2\) &  1.0  \\
    Candidate function coefficient regularization loss \(\mathcal{L}_{reg}\) weight \(\lambda_3\) &  \(5\times10^{-3}\)  \\
    \bottomrule
  \end{tabular}
\end{table}

\subsection{Quartic Oscillator} \label{quartic_oscillator}
\begin{figure}
	\centering
	\includegraphics[width=1.0\linewidth]{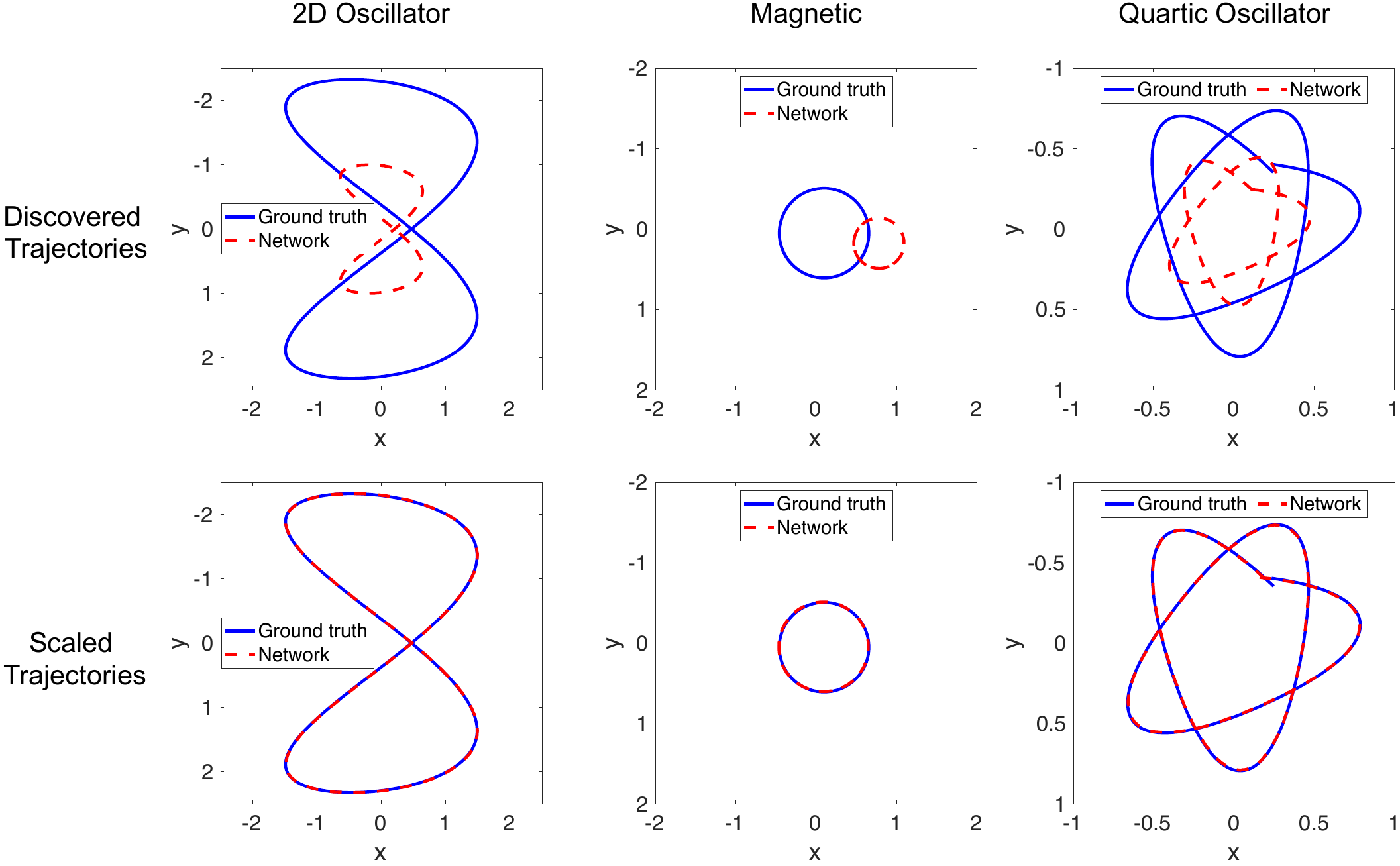}
	\caption{Discovered and scaled (and translated for Magnetic) physical trajectories for second order systems extracted from network.}
	\label{fig:second_order_trajectories}
\end{figure}

The governing equation of Quartic Oscillator is
\begin{equation}
\label{quartic_oscillator_gt}
\begin{pmatrix}\ddot{x}\\\ddot{y}\end{pmatrix}=
\begin{pmatrix}-\mu(x^3+xy^2)\\-\mu(x^2y+y^3)\end{pmatrix}
\end{equation}

where \(\mu=1.0\). With random initial conditions \(x(0),y(0),\dot{x}(0),\dot{y}(0)\in[-0.5,0.5]\), 256 video samples for Quartic Oscillator are generated. Hyperparameters used for training Quartic Oscillator model are shown in Table \ref{hyperparameter_quartic_oscillator}. After 300 iterations of pre-training, the autoencoder reconstruction loss, forward and backward frame reconstruction loss and physical derivative loss are \(3.1\times10^{-6}\), \(5.8\times10^{-6}\) and \(0.14\). After total loss training, the autoencoder reconstruction loss is \(2.8\times10^{-6}\), forward and backward frame reconstruction loss \(2.8\times10^{-6}\) and physical state derivative loss \(0.0040\). For sparse regression coefficients, there are 8 terms having coefficient larger than 0.1. With threshold value 0.1, all redundant terms are filtered out after sequential thresholding. Finally, the network was kept training for 200 epochs with the left 4 candidate terms. The discovered closed-form equation for cubic system is

\begin{equation}
\label{quartic_oscillator_discovery}
\begin{pmatrix}\ddot{x}\\\ddot{y}\end{pmatrix}=
\begin{pmatrix}-2.69x^3-2.63xy^2\\-2.62x^2y-2.69y^3\end{pmatrix}
\end{equation}
After variable transformation with scaling factor 1.65 the discovered Quartic Oscillator system can be rewritten as
\begin{equation}
\label{quartic_oscillator_transformation}
\begin{pmatrix}\ddot{x}\\\ddot{y}\end{pmatrix}=
\begin{pmatrix}-0.99x^3-0.97xy^2\\-0.96x^2y-0.99y^3\end{pmatrix}
\end{equation}
The  comparison of scaled extracted and ground truth physical trajectory is given in Fig \ref{fig:second_order_trajectories}.

\begin{table}
  \caption{Hyperparameter values for the Quartic Oscillator}
  \label{hyperparameter_quartic_oscillator}
  \centering
  \begin{tabular}{lll}
    \toprule
    Parameter  &  Value   \\
    \midrule
    video number  &  256   \\
    video length  &  401  \\
    time step     &  0.05 \\
    \(q\)         &  3 \\
    batch size    &  2    \\
    learning rate &   \(10^{-3}\) \\
    Dynamic model order & 2  \\
    Candidate function polynomial order  &  2   \\
    Physical states derivative loss \(\mathcal{L}_{\dot{\mathbf{x}}_p}\) weight \(\lambda_1\) &  \(1\times10^{-3}\)  \\
    Forward/backward frame reconstruction loss \(\mathcal{L}_{int}\) weight \(\lambda_2\) &  1.0  \\
    Candidate function coefficient regularization loss \(\mathcal{L}_{reg}\) weight \(\lambda_3\) &  \(5\times10^{-3}\)  \\
    \bottomrule
  \end{tabular}
\end{table}

\section{Ablation Studies} \label{ablations}

\begin{figure}
	\centering
	\includegraphics[width=0.90\linewidth]{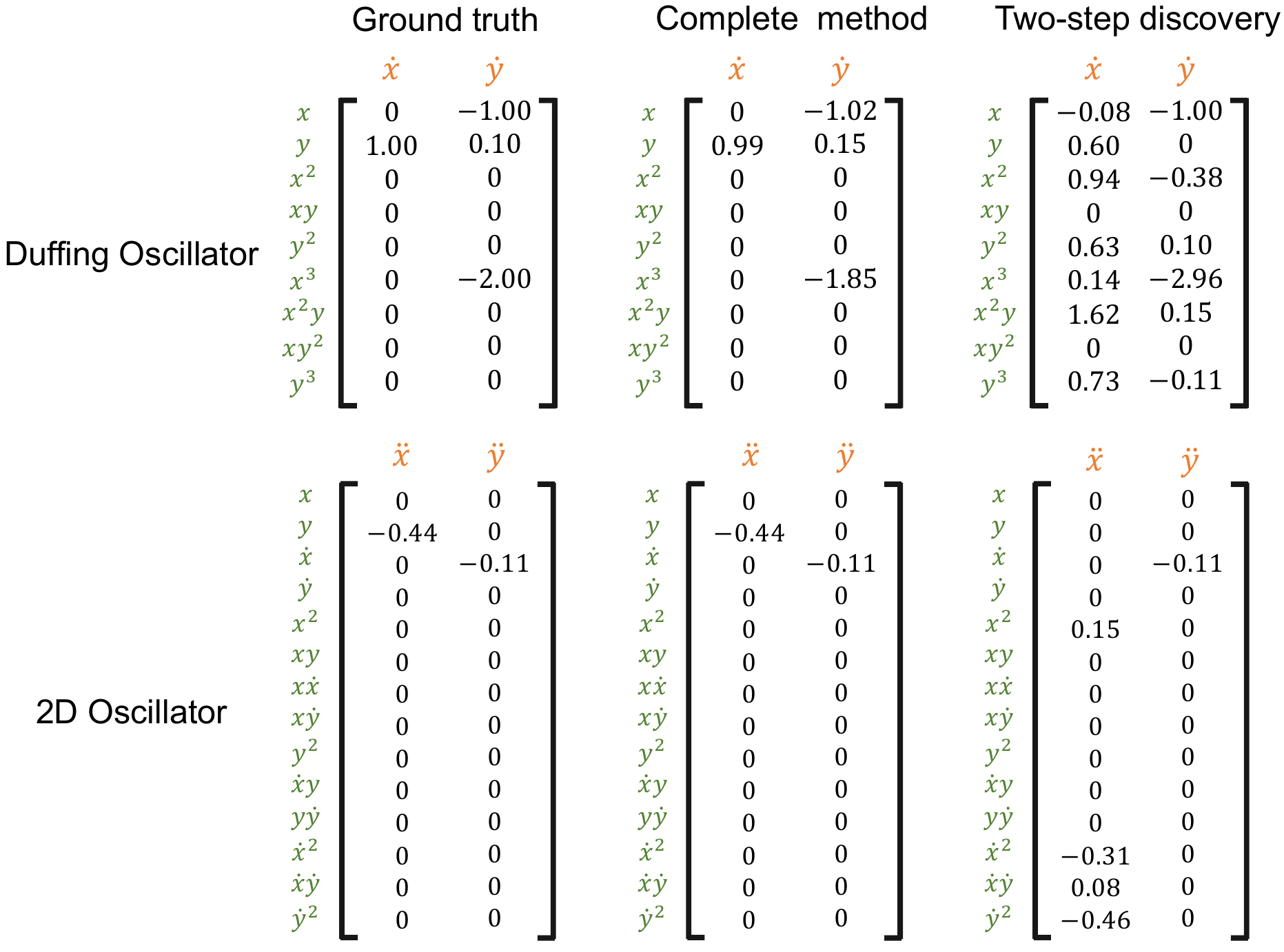}
	\caption{Sparse regression coefficients from two-step discovery for Duffing Oscillator and 2D Oscillator systems.}
	\label{fig:two_step_discovery}
\end{figure}

\begin{figure}
	\centering
	\includegraphics[width=0.90\linewidth]{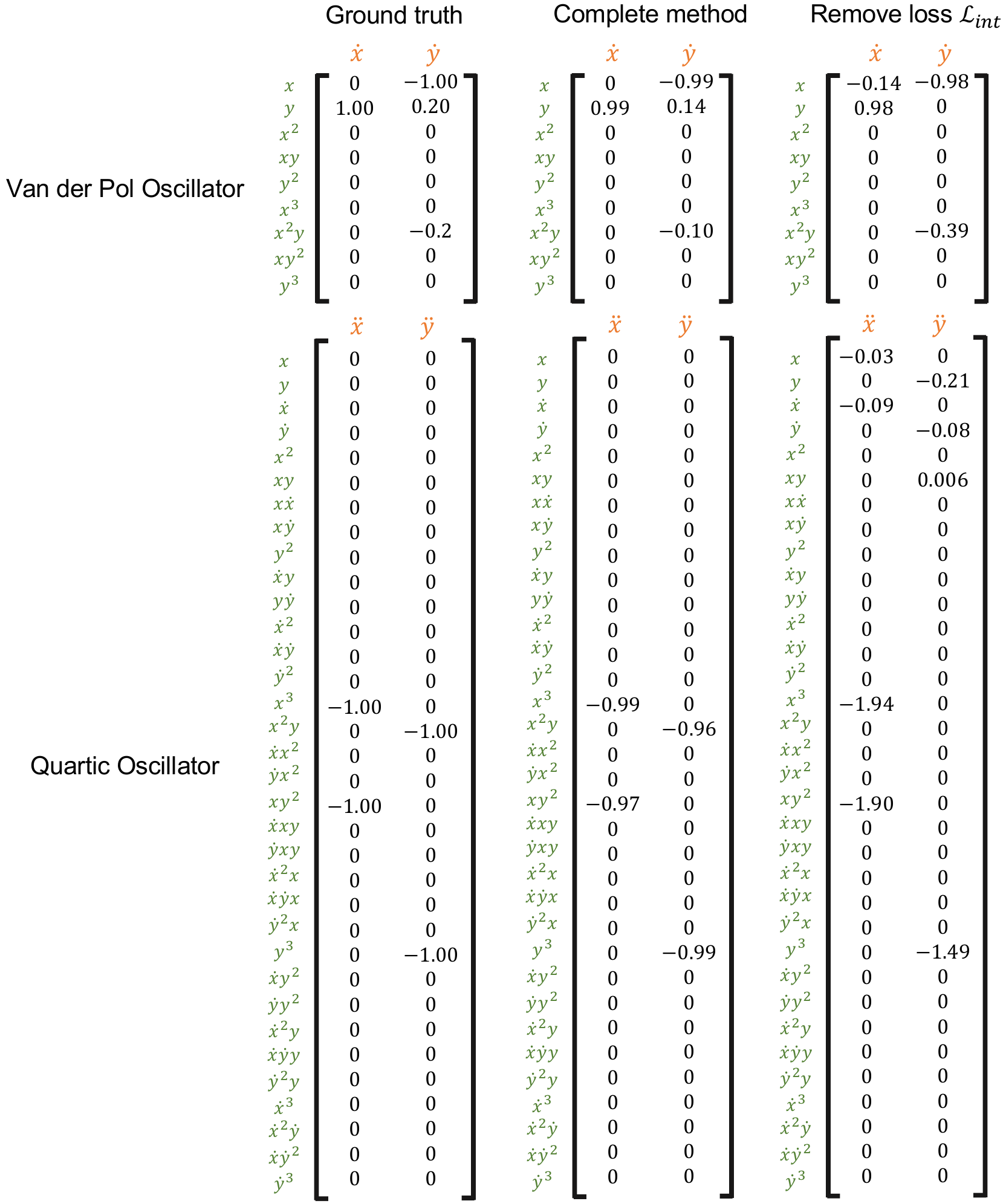}
	\caption{Sparse regression coefficients for Van der Pol and Quartic Oscillator systems in ablation study.}
	\label{fig:physical_constraint_ablation}
\end{figure}

In the two-step discovery, the spatial coordinates of moving object extracted from pre-training model are converted into physical coordinate. Since the real physical coordinate system is unknown, the origin of physical coordinate is set at the centre of video frames. The scaling factor is approximately set as 20, because it will not affect the discovered equation. For the 2D Oscillator, the physical states is obtained as the same procedure as Duffing Oscillator. But the first order derivatives is added to build function library with four variables (physical states and its first order derivative). The parameters of STRidge are as follows: Tolerance value 0.05, \(\lambda=1e-12\), normalization value 1.0. The discovered sparse regression coefficients are given in Fig. \ref{fig:two_step_discovery}. For the ablation study with removing forward and backward frame reconstruction loss from integral physical states, the discovered coefficients for Van der Pol and Quartic Oscillator systems are given in Fig. \ref{fig:physical_constraint_ablation}.

\section{Baseline Comparison} \label{baselines}
\begin{figure}
	\centering
	\includegraphics[width=0.80\linewidth]{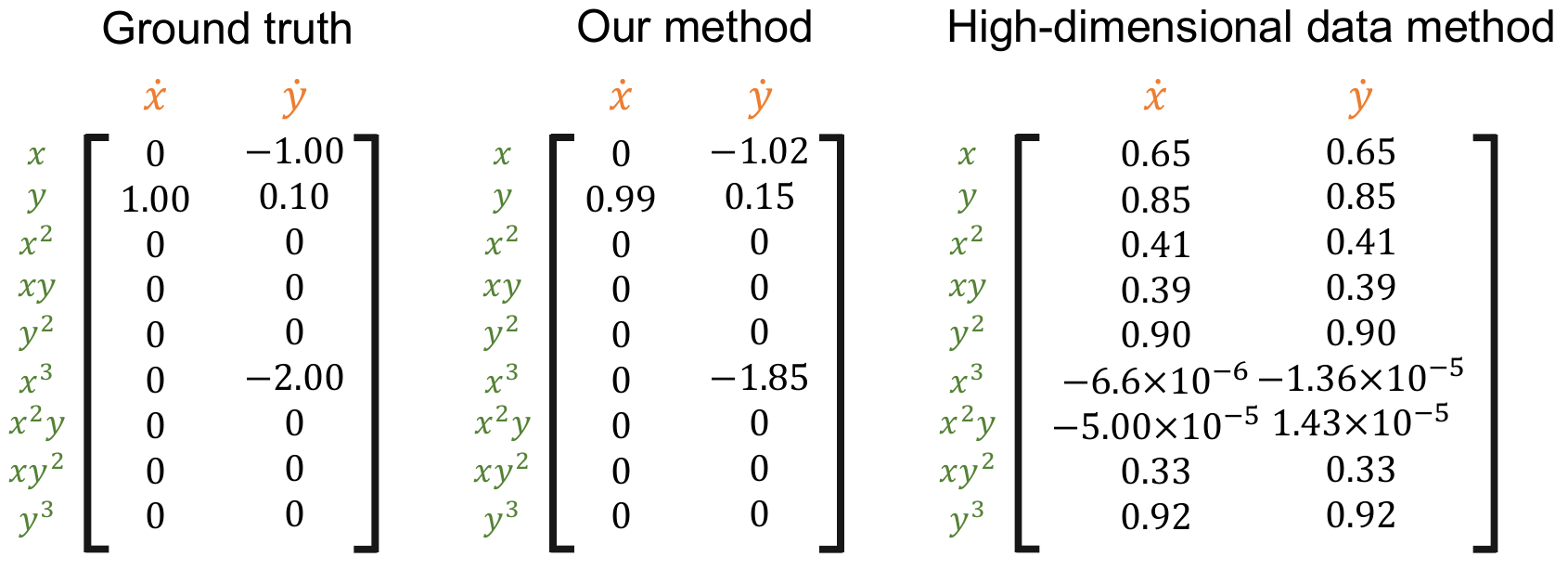}
	\caption{Discovered sparse regression coefficients for Duffing Oscillator by using the method for high-dimensional data discovery proposed in \cite{champion2019data}.}
	\label{fig:baseline_duffing_oscillator}
\end{figure}

Here, the data-driven coordinator and governing equation discovery method for high-dimensional data proposed in \cite{champion2019data} is applied to the video data for discovery. Since this method requires the derivative of input, the RGB video is converted into Gray scale at first. The time derivative of pixel intensity is then calculated by using central difference along temporal direction. Here, the training result for Duffing Oscillator is given in details. Table. \ref{hyperparameter_baseline_duffing} shows the hyperparameters for the discovery. Here, we follow the same way of terming the hyperparameters and the same training procedure given in the paper. During the training, the reconstruction loss and SINDy loss \(\mathcal{L}_{d\mathbf{z}/dt}\) are \(1\times10^{-4}\) and \(2.0\times10^{-6}\), but the SINDy loss \(\mathcal{L}_{d\mathbf{z}/dt}\) does not converge with value of 0.0077. And the discovered SINDy coefficients are given in Fig. \ref{fig:baseline_duffing_oscillator}. It can be seen that the physical coefficients have no sparsity and the discovered model is not correct. This possible reason is that, even though the video can be seen as a high-dimensional function of moving object coordinate, this function is not continuous in time. As a result, the chain rule used in the baseline model to calculate the derivative of latent variable by propagating along the network the time derivatives of input does not hold.

\begin{table}
  \caption{Hyperparameters for Duffing Oscillator within high-dimensional data method}
  \label{hyperparameter_baseline_duffing}
  \centering
  \begin{tabular}{lll}
    \toprule
    Parameter  &  Value   \\
    \midrule
    \(n\)             &  \(64 \times 64\) \\
    \(d\)            &  2   \\
    training samples  &  \(801 \times 64\)  \\
    batch size     &  10000 \\
    activation function         &  sigmoid \\
    Encoder layer widths    &  128,32    \\
    Encoder layer widths    &  32,128    \\
    learning rate &  \(1\times10^{-3}\) \\
    SINDy model order  &  1   \\
    SINDy library polynomial order  & 3\\
    SINDy library includes sine    & no \\
    SINDy loss weight \(\dot{x}\), \(\lambda_1\)   & \(1\times10^{-2}\)\\
    SINDy loss weight \(\dot{z}\), \(\lambda_2\)   & \(1\times10^{-3}\)\\
    SINDy regularization loss weight, \(\lambda_3\)   & \(1\times10^{-5}\)\\
    \bottomrule
  \end{tabular}
\end{table}

\end{document}